\documentclass[pdflatex,sn-mathphys-num, iicol]{sn-jnl}

\usepackage{graphicx}%
\usepackage{multirow}%
\usepackage{amsmath,amssymb,amsfonts}%
\usepackage{amsthm}%
\usepackage{mathrsfs}%
\usepackage[title]{appendix}%
\usepackage{xcolor}%
\usepackage{textcomp}%
\usepackage{manyfoot}%
\usepackage{booktabs}%
\usepackage{algorithm}%
\usepackage{algorithmicx}%
\usepackage{algpseudocode}%
\usepackage{listings}%

\usepackage{colortbl}
\usepackage{diagbox}
\usepackage{subcaption}
\usepackage[capitalize]{cleveref}
\usepackage{bm}

\theoremstyle{thmstyleone}%
\theoremstyle{thmstyletwo}%

\theoremstyle{thmstylethree}%

\begin{document}

\title[Article Title]{MMRL++: Parameter-Efficient and Interaction-Aware Representation Learning for Vision-Language Models}


\author[1]{\fnm{Yuncheng} \sur{Guo}}\email{23210720033@m.fudan.edu.cn}

\author*[1]{\fnm{Xiaodong} \sur{Gu}}\email{xdgu@fudan.edu.cn}

\affil*[1]{\orgdiv{Department of Electronic Engineering}, \orgname{Fudan University}, \orgaddress{\city{Shanghai}, \postcode{200438}, \country{China}}}


\abstract{Large-scale pre-trained \textbf{V}ision-\textbf{L}anguage \textbf{M}odels (VLMs) have significantly advanced transfer learning across a wide range of tasks. However, adapting these models with limited few-shot data often leads to overfitting, undermining their ability to generalize to new tasks. To address this challenge, we propose a novel framework, \textbf{M}ulti-\textbf{M}odal \textbf{R}epresentation \textbf{L}earning (MMRL), which introduces a shared, learnable, and modality-agnostic representation space. Specifically, MMRL generates a set of space tokens, which are projected into both the text and image encoders as representation tokens, facilitating more effective cross-modal interactions. Unlike prior methods that primarily optimize class token features, MMRL integrates representation tokens into the higher layers of the encoders—where task-specific features are more prominent—while preserving general knowledge in the lower layers. During training, both class and representation features are jointly optimized: a trainable projection layer is applied to representation tokens for task adaptation, while the projection layer for class token remains frozen to retain pre-trained knowledge. To further promote generalization, we introduce a regularization term that aligns class and text features with the frozen VLM’s zero-shot features. During inference, we employ a decoupling strategy: both class and representation features are used for base tasks, while only class features, which are more generalizable, are utilized for novel tasks. Building upon this, we propose \textbf{MMRL++}, a parameter-efficient and interaction-aware extension that significantly reduces the number of trainable parameters and enhances intra-modal interactions—particularly across the layers of representation tokens—allowing gradient sharing and instance-specific information to propagate more effectively through the network. Extensive experiments on 15 datasets demonstrate that MMRL and MMRL++ consistently outperform state-of-the-art methods, achieving a strong balance between task-specific adaptation and generalization.}




\maketitle

\section{Introduction}\label{sec1}


\begin{figure}[tb]
\centering
  \includegraphics[width=1.0\linewidth]{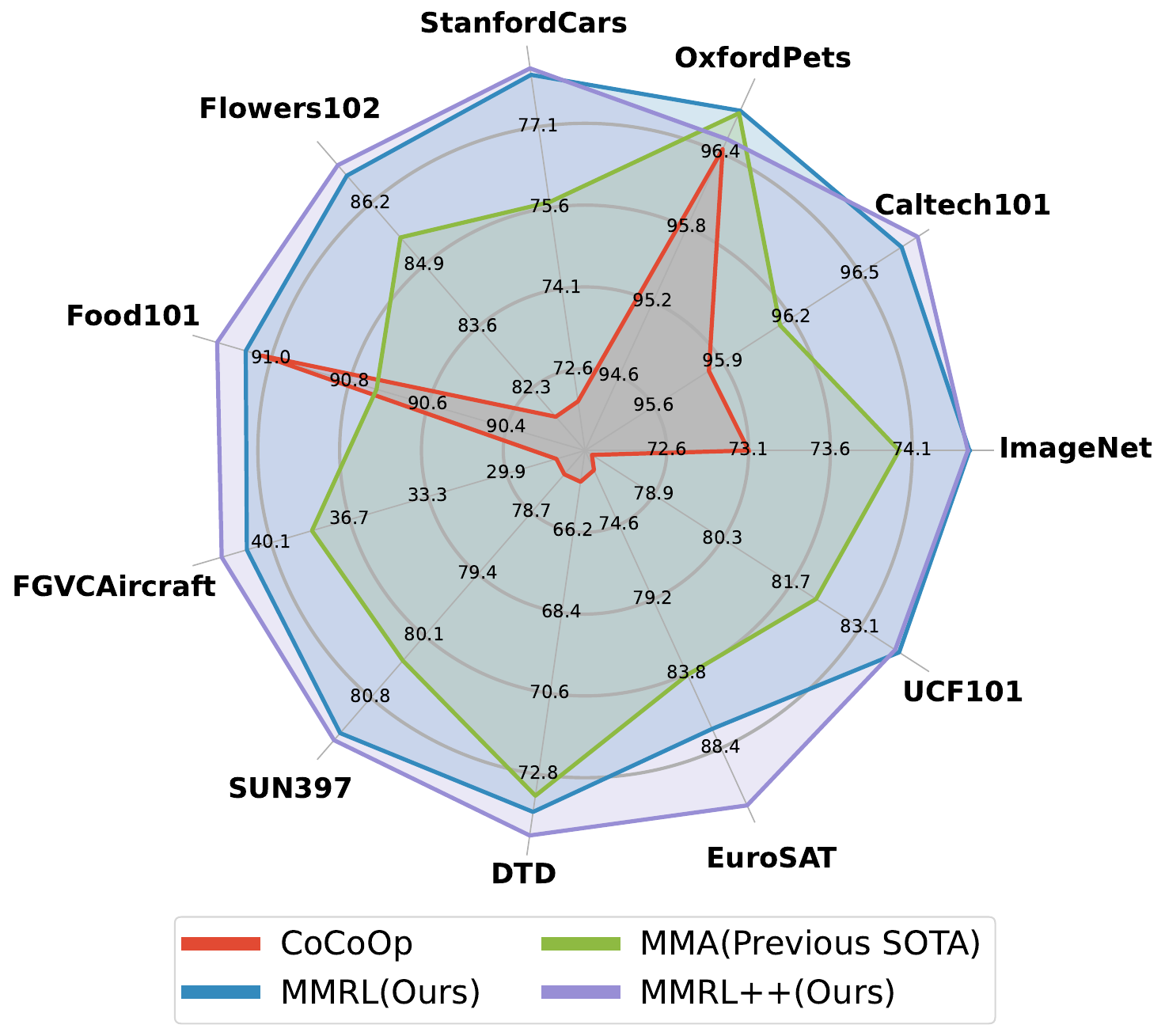}
  \caption{Comprehensive comparison of harmonic mean performance between the baseline CoCoOp, the previous state-of-the-art method MMA, and our proposed MMRL and MMRL++ across 11 diverse datasets for base-to-novel generalization. Our methods demonstrate substantial improvements, consistently outperforming existing approaches.}
  \label{radar}
\end{figure}

Vision-Language Models (VLMs) \cite{clip, align, flamingo, filip, kosmos1, kosmos2, vila} have garnered significant attention for their ability to jointly learn representations from visual and textual data. Among these models, CLIP \cite{clip} stands out as a seminal work, featuring a dual-encoder architecture that employs separate neural networks for processing image and text modalities. Leveraging contrastive learning \cite{contrastive_learning} on a large-scale corpus of over 400 million image-text pairs, CLIP effectively captures complex semantic correspondences across modalities. As a result, it demonstrates strong zero-shot performance on a wide range of tasks, such as medical image analysis \cite{clip_medical1, clip_medical2, clip_medical3}, image and video captioning \cite{clip_captioning1, clip_captioning2, clip_captioning3}, and visual question answering \cite{clip_answering1, clip_answering2, clip_answering3}. Nonetheless, adapting these large-scale pre-trained models to specific downstream tasks remains a non-trivial challenge, primarily due to the high computational cost associated with full-model fine-tuning.

To address this, several techniques have been proposed to enable efficient adaptation of VLMs. Among these, prompt engineering and ensemble-based methods \cite{clip} have shown promise. Prompt engineering involves crafting task-specific textual templates—for instance, using ``A photo of a [CLASS], a type of pet'' when working with the OxfordPets dataset \cite{oxford_pets}. And ensembling techniques enhance robustness by aggregating multiple zero-shot classifiers based on diverse prompts, such as combining ``A photo of a large [CLASS]'' and ``A photo of a small [CLASS]''. Despite their utility, these manually designed strategies often require considerable domain expertise and may not generalize well across tasks or datasets. To mitigate such limitations, CoOp \cite{coop} introduced the concept of continuous prompt learning \cite{prompt_tuning}, in which prompt tokens are represented as learnable vectors while the backbone VLM remains frozen. This strategy allows for task-specific adaptation without the need to fine-tune the entire model. However, the learned context in CoOp does not generalize well to unseen classes within the same dataset, suggesting that it overfits to the base classes observed during training and compromises the generalization capability of the original VLM. Building upon this insight, MaPLe \cite{maple} extended the prompting mechanism by introducing a multimodal prompt learning framework. Rather than relying solely on textual prompts, MaPLe incorporates visual prompts—mapped from textual prompts via a learnable coupling function—into the visual branch, thereby maintaining coherence across modalities. This design has been shown to significantly enhance cross-modal interaction and downstream performance.

\begin{figure}[tb]
\centering
  \includegraphics[width=1.0\linewidth]{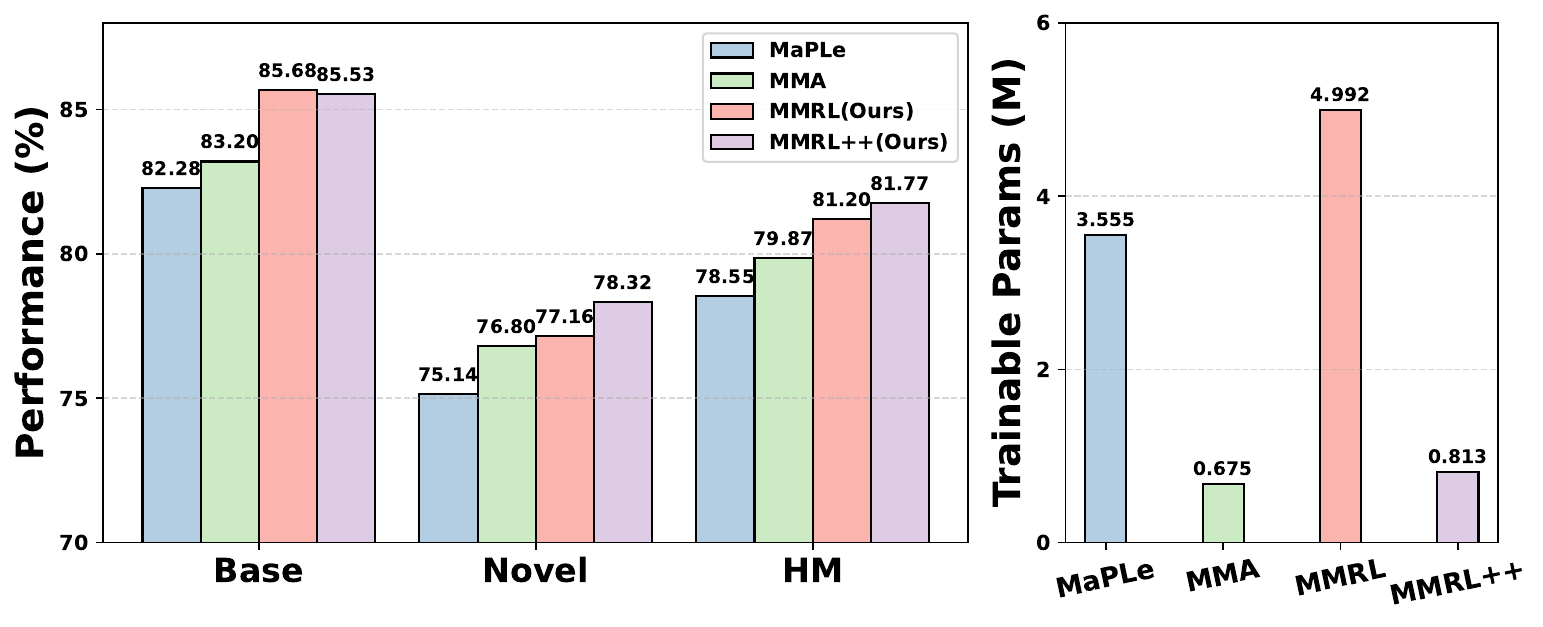}
    \caption{Comprehensive comparison of different efficient multimodal transfer learning approaches (\textit{i.e.}, those incorporating multimodal interaction mechanisms) in terms of performance and the number of trainable parameters.}
  \label{histiogram}
\end{figure}

In parallel, another line of research explores lightweight adaptation modules that refine extracted features without modifying input prompts. For example, CLIP-Adapter \cite{clip-adapter} introduces trainable MLP-based adapters connected to the frozen image encoder via residual pathways, enabling effective feature transformation with minimal computational overhead. Similar to MaPLe, MMA \cite{mma} proposes a multimodal adapter architecture that facilitates cross-branch feature aggregation through a shared projection. Notably, MMA's analysis reveals that different layers within the encoder capture heterogeneous information: lower layers tend to preserve more general representations, while higher layers encode task-specific and discriminative features.

However, existing multimodal deep prompt learning approaches such as MaPLe \cite{maple} typically adopt a shallow-layer prompt concatenation strategy, which may hinder the generalization ability of the model. Moreover, although MaPLe introduces visual prompts derived from textual prompts via a coupling function and incorporates visual information through gradient propagation, its architecture remains inherently text-centric, with optimization primarily focused on textual tokens. This leads to imbalanced multimodal interactions and sub-optimal performance. In addition, both prompt learning and adapter-based methods generally optimize only the class token features using task-specific objectives (\textit{e.g.}, cross-entropy loss). This design choice can intrinsically lead to overfitting, particularly in low-data scenarios such as few-shot learning, thereby compromising the zero-shot and generalization performance of pre-trained VLMs.

To address these issues, we propose a novel framework, termed \textbf{M}ulti-\textbf{M}odal \textbf{R}epresentation \textbf{L}earning (\textbf{MMRL}), which diverges from conventional prompt- and adapter-centric paradigms. At its core, MMRL introduces a shared, modality-agnostic representation space within the upper layers of the vision and language encoders. This space serves as a unified interaction bridge, from which learnable space tokens are projected into both the visual and textual domains through representation aligners. These projected tokens are then concatenated with the original encoder tokens as representation tokens, enabling enhanced cross-modal fusion and alignment. Our representation tokens are designed to capture dataset-specific knowledge for downstream tasks, while the original class token is regularized to retain generalizable knowledge embedded in the pre-trained VLM. MMRL offers three key advantages: (1) an unbiased shared representation space that fosters balanced and efficient multimodal learning; (2) preservation of the original VLM’s generalization ability by avoiding prompt integration at shallow encoder layers; and (3) a decoupled inference mechanism, unlike prompt learning and adapter-based methods that refine only the class token features. During training, we prioritize optimizing representation token features, with their projection layers being trainable, while keeping the original class token’s projection layer fixed. To further preserve the generalizability of the class token, a regularization term aligns its features with the zero-shot features from the frozen VLM. For inference, both representation and class token features are utilized for base tasks, whereas only the class token is employed for unseen classes or new tasks.

\begin{figure*}[tb]
\centering
  \includegraphics[width=1.0\linewidth]{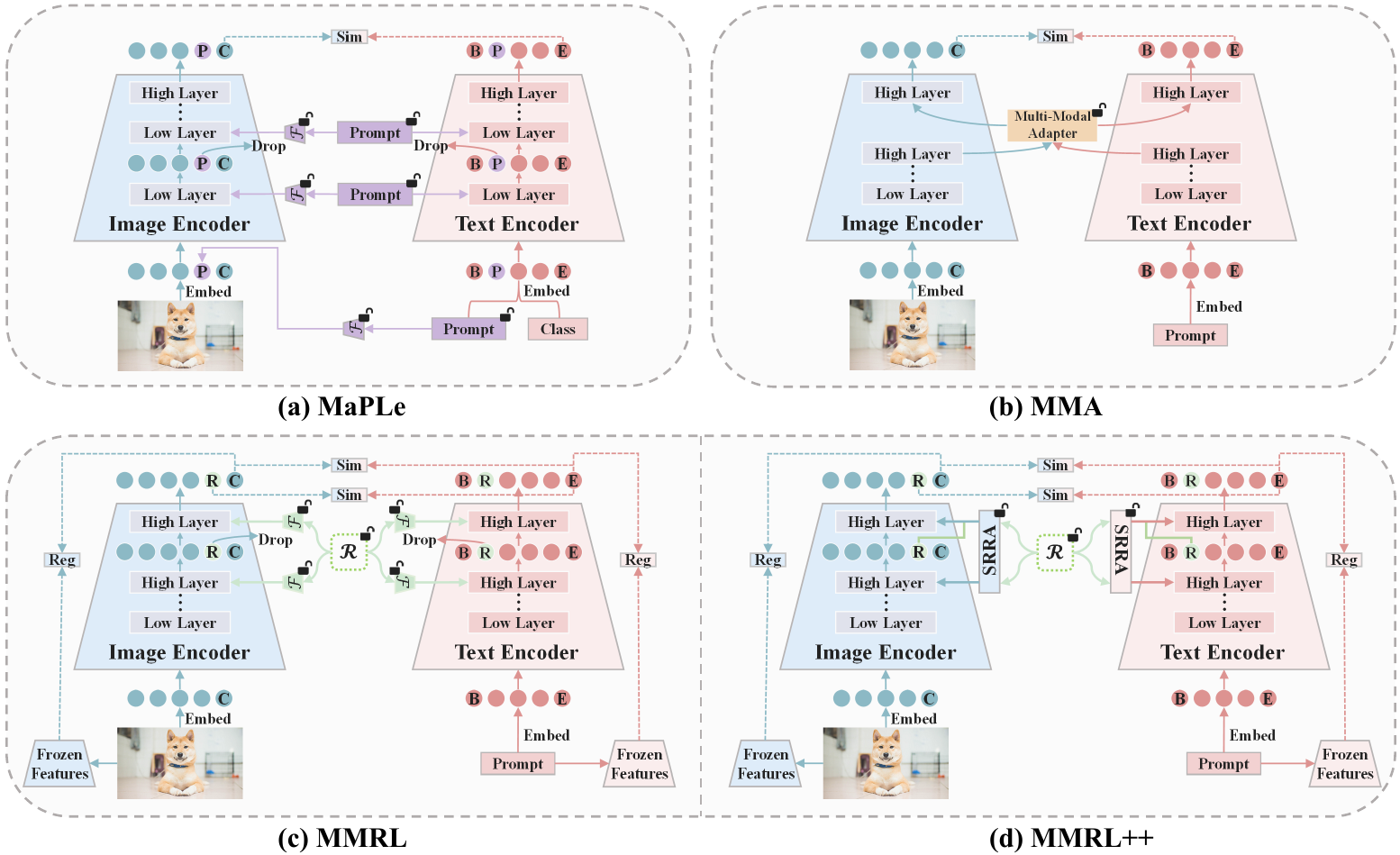}
  \caption{Comparison of the proposed MMRL and MMRL++ frameworks with representative efficient multimodal transfer learning approaches—MaPLe from prompt learning and MMA from adapter-style learning.}
  \label{framework_comparison}
\end{figure*}

Despite achieving state-of-the-art performance, MMRL has certain limitations. Specifically, it introduces a substantial number of learnable parameters due to the extensive representation aligners per layer, which may restrict its practical applicability and increase the risk of overfitting. Furthermore, the representation tokens in MMRL are propagated in isolation across different layers, potentially leading to sub-optimal performance and training instability. Since the feature projection layer of the representation tokens is learnable, this layer-wise isolation can introduce significant perturbations to the pretrained model’s functionality, further amplifying the risk of overfitting.

To mitigate these limitations, we propose \textbf{MMRL++}, an enhanced version of our MMRL \footnote{This paper extends our earlier work MMRL accepted by The IEEE/CVF Conference on Computer Vision and Pattern Recognition 2025 (CVPR2025).} \cite{mmrl} framework. Drawing inspiration from recent LoRA-related research \cite{lora, mole, ders}, we decompose multiple representation aligners into a shared representation aligner with multiple low-rank residual aligners. This approach offers two key benefits: first, by shared representation aligners, we enhance gradient interactions across different layers of representation tokens, and second, we significantly reduce the number of learnable parameters, resulting in a more lightweight adaptation strategy with a lower risk of overfitting. Additionally, we introduce a progressive representation composition mechanism, which enables deep representation tokens to fully assimilate and retain instance-level information across different layers, further improving training stability and generalization.

Our primary contributions are summarized as follows.
\begin{itemize}
\item We propose \textbf{MMRL}, a novel multimodal representation learning framework that establishes a shared, unbiased, learnable representation space in the upper layers of vision and language encoders. This design facilitates fine-grained cross-modal alignment while maintaining the structural integrity of pre-trained models.
\item To preserve the generalization capacity of VLMs, we introduce a decoupled adaptation strategy, where task-specific representation tokens are learned independently from the original class token. The class token is regularized to retain zero-shot capabilities, enabling both transferability and generalizability.
\item We further develop \textbf{MMRL++}, an enhanced variant featuring two key designs: (1) a \textbf{Shared-Residual Representation Aligner (SRRA)} framework, which reduces trainable parameters via low-rank decomposition and shared projections across layers, while simultaneously enabling implicit gradient interaction and knowledge sharing during training; and (2) a \textbf{Progressive Representation Composition (PRC)} mechanism that facilitates inter-layer semantic flow, improving learning stability and representation generalization.
\item Extensive experiments on various downstream benchmarks demonstrate that MMRL and MMRL++ outperform existing prompt learning and adapter-style learning methods, achieving superior performance in both adaptation effectiveness and generalization.
\end{itemize}

\section{Related Work}
\subsection{Vision-Language Models}
Vision-language models (VLMs) have emerged as a powerful paradigm for learning unified multimodal representations, surpassing traditional architectures that rely exclusively on either visual or textual supervision. Recent breakthroughs in this domain, such as CLIP~\cite{clip}, ALIGN~\cite{align}, FILIP~\cite{filip}, KOSMOS~\cite{kosmos1, kosmos2}, and VILA~\cite{vila}, have demonstrated impressive performance across a wide array of downstream tasks. These models are typically trained in a self-supervised manner on massive datasets comprising image-text pairs, enabling them to capture semantic correspondences across modalities. For example, CLIP is trained on 400 million image-text pairs, whereas ALIGN utilizes an even larger dataset of 1.8 billion pairs. While such large-scale pre-training confers remarkable generalization capabilities, adapting these models effectively to specific downstream tasks remains an open challenge due to high computational cost.

\subsection{Efficient Transfer Learning}
To address the adaptation bottleneck of pre-trained VLMs, a variety of efficient transfer learning techniques have been proposed, prominently including prompt learning and adapter-style learning strategies.

Prompt learning \cite{prompt_tuning, prefix_tuning, p_tuning} has shown notable success in tuning VLMs with minimal parameter updates. CoOp~\cite{coop} initiates this line of work by introducing learnable continuous vectors to replace handcrafted textual templates. Although this enhances flexibility, it often undermines the original model’s zero-shot and generalization performance. CoCoOp~\cite{cocoop} improves upon this by generating instance-specific prompts using visual cues, thereby enhancing robustness to distribution shifts. ProDA~\cite{proda} explores prompt distribution learning to foster better adaptability, while PLOT~\cite{plot} aligns vision and language features using an optimal transport framework. Other methods emphasize preserving general textual knowledge: KgCoOp~\cite{kgcoop} minimizes divergence between learned and manual prompts, and ProGrad~\cite{prograd} restricts gradient updates to those consistent with general knowledge. RPO~\cite{rpo} addresses internal representation shifts via masked attention mechanisms. Beyond textual prompting, methods such as MaPLe~\cite{maple} incorporate visual prompts derived from textual prompts via a learnable coupling function, enabling tighter cross-modal alignment. ProVP \cite{provp} employs single-modal visual prompts with contrastive feature re-formation to align prompted visual features with CLIP's distribution. PromptSRC~\cite{promptsrc} utilizes self-regularization to prevent overfitting, while MetaPrompt~\cite{metaprompt} applies a meta-learning-based tuning strategy to produce domain-generalizable prompts. TCP~\cite{tcp} enhances generalization capabilities by converting textual knowledge into class-aware tokens.

Adapter-style learning methods offer another lightweight and effective alternative for fine-tuning VLMs. CLIP-Adapter~\cite{clip-adapter} integrates small-scale two-layer MLP adapters after the image encoder to adjust CLIP’s feature space via a cross-entropy loss. Tip-Adapter~\cite{tip-adapter} further enhances efficiency by caching training features for similarity computation during inference. However, these methods treat the image and text branches independently, potentially limiting cross-modal alignment. To address this, MMA~\cite{mma} constructs a shared projection between the adapters of both modalities, linking the visual and textual branches to enable cross-modal gradient propagation and foster better alignment.

In addition to the aforementioned approaches, several recent works~\cite{coprompt, hpt, argue, lweib, promptkd} leverage large language models (LLMs), such as GPT-3~\cite{gpt3}, to augment textual prompts or employ dataset-wide distillation to enhance performance. While these methods have shown promise, their substantial computational overhead may places them beyond the intended scope of efficient transfer learning.

\begin{figure*}
\centering
  \includegraphics[width=1.0\linewidth]{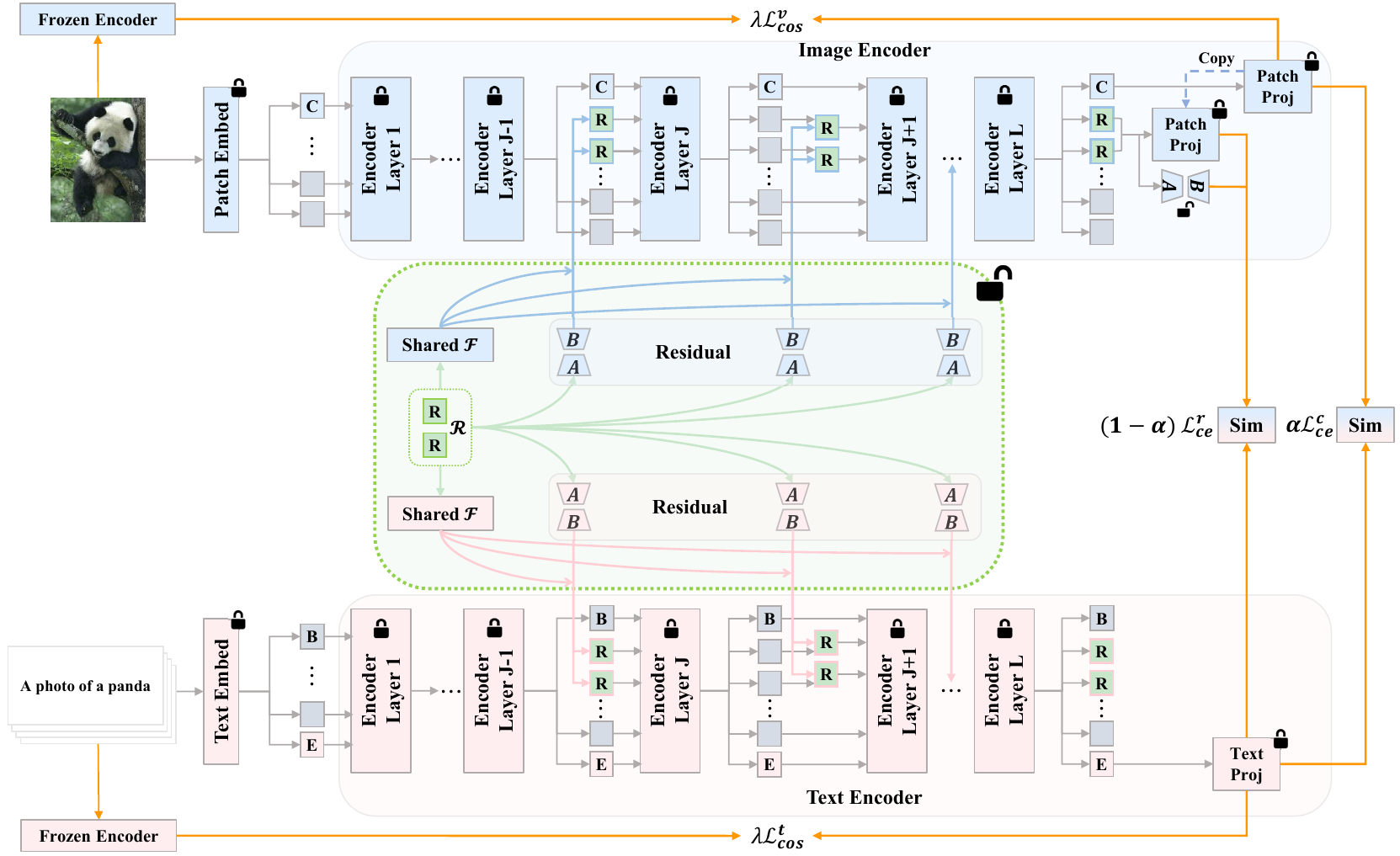}
  \caption{Overview of the MMRL++ training pipeline. `C' denotes the class token, `B' the BOS token, `E' the EOS token, `$\mathcal{R}$' our shared representation space, and `R' the representation tokens. The components subject to optimization include the representation space $\mathcal{R}$, the shared representation aligner $\mathcal{F}$, residual representation aligners ($A$, $B$), and the residual patch projection head for representation tokens, while the pre-trained CLIP model remains entirely frozen. To preserve generalization knowledge, representation tokens are integrated into both encoders starting from layer $J$.}
  \label{framework1}
\end{figure*}

\section{Method}
\label{sec:method}
Our approach, in line with previous methods, builds upon a pre-trained VLM, CLIP \cite{clip}. In this section, we present the proposed \textbf{MMRL} and \textbf{MMRL++} frameworks in detail.

\subsection{Preliminary}
We begin by introducing the notations and backbone components of our approach. CLIP consists of two primary modules: an image encoder $\mathcal{V}$ and a text encoder $\mathcal{W}$.

\noindent \textbf{Image Encoding:} The image encoder $\mathcal{V}$ consists of $L$ Transformer~\cite{transformer} layers, denoted as $\{\mathcal{V}_i\}_{i=1}^{L}$. Given an input image $x \in \mathbb{R}^{H \times W \times 3}$, it is partitioned into $M$ fixed-size patches and projected into patch embeddings, yielding an initial embedding matrix $E_0 \in \mathbb{R}^{M \times d_v}$, where $d_v$ denotes the embedding dimension. A learnable class token $c_0$ is prepended to $E_0$, and positional embeddings are added. The combined sequence is iteratively processed through the Transformer layers,
\begin{equation}
    [c_i, E_i] = \mathcal{V}_i([c_{i-1}, E_{i-1}]), \quad i = 1, 2, \ldots, L
\end{equation}
After the final layer, the output class token $c_L$ is projected into a shared vision-language latent space via a patch projection head $P_v^c$,
\begin{equation}
    f = P_v^c(c_L)
\end{equation}
where $f \in \mathbb{R}^d$ represents the final image feature.

\noindent \textbf{Text Encoding:} For an input text, \textit{e.g.}, ``A photo of a [CLASS].", it is tokenized and converted into embeddings $T_0 \in \mathbb{R}^{N \times d_t}$, where $N$ denotes the token length and $d_t$ the embedding dimension. Beginning-of-sentence (BOS) and end-of-sentence (EOS) tokens, denoted as $b_0$ and $e_0$, are appended to mark the sequence boundaries. These token embeddings, along with positional encodings, are passed through the text encoder's $L$ Transformer layers, $\{\mathcal{W}_i\}_{i=1}^{L}$, as follows,
\begin{equation} 
    [b_i, T_i, e_i] = \mathcal{W}_i([b_{i-1}, T_{i-1}, e_{i-1}]) \quad i = 1, \cdots, L 
\end{equation}
After the final layer, the output of the EOS token, $e_L$, is projected into the shared vision-language space via a text projection layer $P_t$,
\begin{equation} 
    w = P_{t}(e_{L})
\end{equation} 
where $w \in \mathbb{R}^d$ denotes the final text feature.

\noindent \textbf{Classification with CLIP:} With the image feature $f$ and text features $\{w_c\}_{c=1}^C$ for $C$ classes, CLIP calculates the cosine similarity between $f$ and each $w_c$,
\begin{equation} 
    \text{sim}(f, w_c) = \frac{f \cdot w_c}{\|f\| \, \|w_c\|}
\end{equation}
where $\|\cdot\|$ represents the $L_2$ norm. Class probabilities are then obtained using the softmax function,
\begin{equation} 
    p(y = c \mid f) = \frac{\exp(\text{sim}(f, w_c) / \tau)}{\sum_{i=1}^{C} \exp(\text{sim}(f, w_i) / \tau)}
\end{equation} 
where $\tau$ is a temperature parameter. The final predicted class is selected as the one with the highest probability score.

\subsection{Multi-Modal Representation Learning (MMRL)} Our proposed MMRL framework aims to address the challenge of adapting pre-trained VLMs with limited few-shot data while preserving their ability to generalize to novel tasks. The training and inference pipelines for MMRL and its extension MMRL++ are illustrated in \cref{framework1} and \cref{framework2}, respectively. Below, we detail the core components of the methodology.

\subsubsection{Learnable Representation Space}
MMRL establishes a shared, learnable representation space $\mathcal{R}$ to facilitate multimodal interactions, initialized through sampling from a Gaussian distribution. A learnable representation aligner $\mathcal{F}(\cdot)$, implemented as a linear projection, transforms the space tokens $R \in \mathbb{R}^{K \times d_r}$—where $K$ is the number of tokens and $d_r$ the dimension of the representation space—into visual and textual modality-specific forms,
\begin{align}
    R^v &= \{R_i^v\}_{i=J-1}^{L-1} \quad where & R_i^v = \mathcal{F}_i^v(R) \\ 
    R^t &= \{R_i^t\}_{i=J-1}^{L-1} \quad where & R_i^t = \mathcal{F}_i^t(R)
\end{align}
Here, $R_i^v \in \mathbb{R}^{K \times d_v}$ and $R_i^t \in \mathbb{R}^{K \times d_t}$ denote the visual and textual representation tokens injected into the $(i+1)$-th Transformer layer. The index $J$ indicates the starting layer for token integration.

\subsubsection{Integration into Upper Encoder Layers}
To preserve the general knowledge encoded in the lower layers of the pre-trained CLIP model, we insert the representation tokens $\mathcal{R}^v$ and $\mathcal{R}^t$ only into the upper layers of the image encoder $\mathcal{V}$ and text encoder $\mathcal{W}$, beginning from layer $J$.

For the image encoder $\mathcal{V}$, the input sequence is processed as follows,
\begin{align}
    [c_i, E_i] &= \mathcal{V}_i([c_{i-1}, E_{i-1}]) \nonumber \\ 
    &\hspace{2cm} i = 1, \ldots, J-1 \\  
    [c_i, \_, E_i] &= \mathcal{V}_i([c_{i-1}, R_{i-1}^v, E_{i-1}]) \nonumber \\ 
    &\hspace{2cm} i = J, \ldots, L - 1 \\
    [c_i, R_i^v, E_i] &= \mathcal{V}_i([c_{i-1}, R_{i-1}^v, E_{i-1}]) \quad i = L \label{R_L^v}
\end{align}

For the text encoder $\mathcal{W}$, unlike previous prompt learning method MaPLe \cite{maple} that replaces part of the tokens of $T_i$ with deep prompts, we retain the entire $T_i$ and insert $R_i^t$ before it, aiming to preserve the original textual information,
\begin{align}
    [b_i, T_i, e_i] &= \mathcal{W}_i([b_{i-1}, T_{i-1}, e_{i-1}]) \quad \nonumber \\ 
    &\hspace{2cm} i = 1, \ldots, J-1  \\  
    [b_i, \_, T_i, e_i] &= \mathcal{W}_i([b_{i-1}, R_{i-1}^t, T_{i-1}, e_{i-1}]) \nonumber \\
    &\hspace{2cm} i = J, \ldots, L-1  \\
    [b_i, R_i^t, T_i, e_i] &= \mathcal{W}_i([b_{i-1}, R_{i-1}^t, T_{i-1}, e_{i-1}]) \nonumber \\ 
    &\hspace{2cm} \quad i = L \label{R_L^t}
\end{align}
Note that due to the autoregressive nature of the text encoder, we modify the attention mask matrix to accommodate the extended sequence length.

\subsubsection{Representation Learning}
Our goal is to enable dataset-specific adaptation via representation tokens while preserving the pre-trained knowledge of the class token. Through a set of strategies aimed at retaining generalization during both training and inference, MMRL enables flexible inference for different tasks, as detailed below.

\begin{itemize} 

\item \textbf{Training Phase:} We optimize the features of both the representation tokens and the original class token, with the primary focus on representation features to preserve pre-trained knowledge. Specifically, the projection layer for the representation tokens is trainable, while that for the class token remains fixed. For the image encoder $\mathcal{V}$, after passing through $L$ Transformer layers, we obtain the output $c_L \in \mathbb{R}^{d_v}$ for the class token and $R_L^v \in \mathbb{R}^{K \times d_v}$ for the $K$ representation tokens. The aggregated output of the representation tokens, $r_L$, is derived by mean pooling across the $K$ tokens,
\begin{equation}
    r_L = \text{Mean}(R_L^v)
\end{equation}
where $r_L \in \mathbb{R}^{d_v}$. We then project these into the common V-L latent space, yielding the class features $f_c$ and representation features $f_r$.
\begin{equation}
    f_c = P_v^c(c_L) \quad f_r = P_v^r(r_L)
\end{equation}
Here, $P_v^c$ is the original, frozen patch projection layer of CLIP for class features, while $P_v^r$ for representation features is trainable.

For the text encoder $\mathcal{W}$, following the sequential nature of text, we map the EOS token $e_L$—as in the original CLIP model—after processing through $L$ transformer layers into the common V-L space, yielding the text features.
\begin{equation} 
    w = P_{t}(e_{L})
\end{equation}
With the image features $f_c$, $f_r$, and the text classifiers $\{w_c\}_{c=1}^C$ for $C$ classes, we apply cross-entropy loss to separately optimize the class and representation features,
\begin{align}
\setlength\abovedisplayskip{3pt}
\setlength\belowdisplayskip{3pt}
    \mathcal{L}_{ce}^c &= -\sum_c^C y_c \log p(y = c \mid f_c) \\
    \mathcal{L}_{ce}^r &= -\sum_c^C y_c \log p(y = c \mid f_r)
\end{align}
where $y_c = 1$ if the image $x$ belongs to class $c$, and $y_c = 0$ otherwise. 

To further preserve the generalization of class features, we enforce consistency between $(f_c, w)$ and the frozen CLIP features $(f_0, w_0)$ via cosine similarity, explicitly guiding the training trajectory,
\begin{align}
    \mathcal{L}_{cos}^v &= 1 - \frac{f_c \cdot f_0}{\|f_c\| \, \|f_0\|} \\
    \mathcal{L}_{cos}^t &= 1 - \frac{1}{C}\sum_c^C \frac{w^c \cdot w_0^c}{\|w^c\| \, \|w_0^c\|}
\end{align}
The overall loss function is
\begin{equation}
    \mathcal{L}_{MMRL} = \alpha \mathcal{L}_{ce}^c + (1 - \alpha) \mathcal{L}_{ce}^r + \lambda (\mathcal{L}_{cos}^v + \mathcal{L}_{cos}^t)
\end{equation}
where $\alpha$ controls the balance between the features, and $\lambda$ is the penalty coefficient.

\begin{figure}[tb]
\centering
  \includegraphics[width=1.0\linewidth]{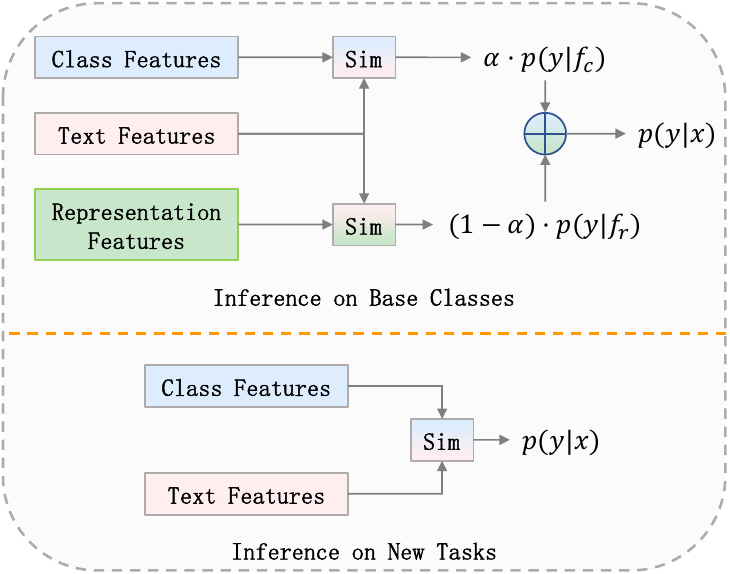}
  \caption{Inference pipeline of MMRL and MMRL++, where feature selection dynamically adjusts to in-distribution and out-of-distribution tasks for improved generalization and flexibility.}
  \label{framework2}
\end{figure}

\item \textbf{Testing on Base Classes:} 
For in-distribution classes seen during training, we combine the generalizable class token features and the dataset-adaptive representation features. The probability of an in-distribution test sample $x$ belonging to the $c$-th class is
\begin{align}
    p(y = c \mid x) &= \alpha \cdot p(y = c \mid f_c) \nonumber \\ 
    &\hspace{0.5cm} + (1-\alpha) \cdot p(y = c \mid f_r)
\end{align}
where $f_c$ and $f_r$ are features extracted from the class token and representation tokens, respectively.

\item  \textbf{Testing on Novel Classes:}
For unseen classes or new tasks, we rely solely on the generalized features of the class token.
\begin{equation}
    p(y = c \mid x) = p(y = c \mid f_c)
\end{equation}
\end{itemize}

\subsection{MMRL++}
Building upon the foundation of MMRL, we identify and address two critical limitations that hinder its scalability and stability. First, the use of multiple independent representation aligners between representation space and encoder layers significantly increases the number of trainable parameters. This not only undermines the lightweight design of the framework but also heightens the risk of overfitting. Second, the absence of interactions among representation tokens across layers leads to isolated learning, which can disrupt the behavior of the pre-trained model, leading to unstable training dynamics and further increasing the susceptibility to overfitting.

To overcome these challenges, we introduce two key enhancements in \textbf{MMRL++}: (1) \textbf{Shared-Residual Representation Aligner}, which significantly reduces parameter overhead while preserving layer-wise expressivity and facilitating gradient sharing between layers; and a \textbf{Progressive Representation Composition} mechanism, which enables inter-layer token interactions to deliver instance-specific features during adaptation. We elaborate on each component below.

\subsubsection{Shared-Residual Representation Aligner (SRRA)}
In our previous architecture, the representation tokens at each high layer are projected from the representation space via different independent representation aligners per layer. This approach not only results in a significant proliferation of trainable parameters but also incurs potential overfitting risks. To overcome these limitations, we introduce the \emph{Shared-Residual Representation Aligner (SRRA)}, a parameter-efficient and structure-aware alternative that disentangles shared and specialized knowledge within the aligner modules.

SRRA factorizes the aligner weight at each layer into two distinct components: a globally shared part and a layer-specific residual. Specifically, for each of the $(L-J+1)$ aligners associated with a particular modality $m \in \{v, t\}$, the trained aligner weight matrix can be decomposed as:
\begin{equation}
W_i^m = W_{init}^m + \Delta W_i^m \quad  i = J-1, \cdots, L-1
\end{equation}
where $W_{init}^m \in \mathbb{R}^{d_r \times d_m}$ denotes the initialized weights, and $\Delta W_i^m \in \mathbb{R}^{d_r \times d_m}$ captures the layer-specific residual knowledge learned during training. Since all aligners are initialized identically in practice, we can treat $W_{init}$ as a globally shared component across layers, denoted as $W_{shared}$. Thus, the final form becomes:
\begin{equation}
W_i^m = W_{shared}^m + \Delta W_i^m \quad  i = J-1, \ldots, L-1
\end{equation}
With this decomposition, inspired by LoRA~\cite{lora}, we model each residual $\Delta W_i$ as a low-rank decomposition, significantly reducing the number of trainable parameters:
\begin{equation}
W_i^m = W_{shared}^m + A_i^mB_i^m \quad  i = J-1, \cdots, L-1
\end{equation}
where $A_i \in \mathbb{R}^{d_r \times r_1}$ and $B_i \in \mathbb{R}^{r_1 \times d_m}$, and $r_1$ is the dimension of the low-rank space in LoRA. Notably, only a single shared weight $W_{shared}$ is maintained for the entire aligner stack, enabling extensive weight sharing and alleviating the redundancy caused by independent aligners. This design empowers the model to retain strong layer-wise expressiveness while introducing minimal additional parameters. Moreover, SRRA promotes stable training and better generalization by enforcing a shared representation aligner across layers.

Similarly, for the patch projection layer $P_v^r$ of representation tokens, we define its weight as:
\begin{equation}
    W_r = W_v^c + A_rB_r
\end{equation}
where $W_v^c \in \mathbb{R}^{d_v \times d}$ is the frozen weight of the original CLIP patch projection layer $P_v^c$ for the class token, and $A_r \in \mathbb{R}^{d_v \times r_2}$ and $B_r \in \mathbb{R}^{r_2 \times d}$ are the learnable low-rank residual weight for the representation tokens.

\begin{table*}[t]
\centering
\caption{Summary of the 15 datasets.}
\label{datasets}
\renewcommand\arraystretch{1.1}
\resizebox{1.0\textwidth}{!}{
    \begin{tabular}{@{}l|llllll@{}}
    \toprule
    Dataset      & Classes & Train  & Val    & Test   & Description                         & Prompt                                \\ \midrule
    ImageNet     & 1000    & 1.28M  & $\sim$ & 50000  & Recognition of generic objects      & ``a photo of a [CLASS].”              \\
    Caltech101   & 100     & 4128   & 1649   & 2465   & Recognition of generic objects      & ``a photo of a [CLASS].”              \\
    OxfordPets      & 37    & 2944   & 736    & 3669   & Fine-grained classification of pets                    & ``a photo of a [CLASS], a type of pet.”      \\
    StanfordCars & 196     & 6509   & 1635   & 8041   & Fine-grained classification of cars & ``a photo of a [CLASS].”              \\
    Flowers102      & 102   & 4093   & 1633   & 2463   & Fine-grained classification of flowers                 & ``a photo of a [CLASS], a type of flower.”   \\
    Food101         & 101   & 50500  & 20200  & 30300  & Fine-grained classification of foods                   & ``a photo of [CLASS], a type of food.”       \\
    FGVCAircraft    & 100   & 3334   & 3333   & 3333   & Fine-grained classification of aircrafts               & ``a photo of a [CLASS], a type of aircraft.” \\
    SUN397       & 397     & 15880  & 3970   & 19850  & Scene classification                & ``a photo of a [CLASS].”              \\
    DTD          & 47      & 2820   & 1128   & 1692   & Texture classification              & ``[CLASS] texture.”                   \\
    EuroSAT         & 10    & 13500  & 5400   & 8100   & Land use \& cover classification with satellite images & ``a centered satellite photo of [CLASS].”    \\
    UCF101       & 101     & 7639   & 1898   & 3783   & Action recognition                  & ``a photo of a person doing [CLASS].” \\ \midrule
    ImageNetV2   & 1,000   & $\sim$ & $\sim$ & 10,000 & New test data for ImageNet          & ``a photo of a [CLASS].”              \\
    ImageNet-Sketch & 1,000 & $\sim$ & $\sim$ & 50,889 & Sketch-style images of ImageNet classes                & ``a photo of a [CLASS].”                     \\
    ImageNet-A      & 200   & $\sim$ & $\sim$ & 7,500  & Natural adversarial examples of 200 ImageNet classes   & ``a photo of a [CLASS].”                     \\
    ImageNet-R   & 200     & $\sim$ & $\sim$ & 30,000 & Renditions of 200 ImageNet classes  & ``a photo of a [CLASS].”              \\ \bottomrule
    \end{tabular}
    }
\end{table*}

\subsubsection{Progressive Representation Composition (PRC)}
While the proposed \emph{SRRA} introduces shared aligner across layers—effectively mitigating the problem of entirely independent token learning—it does not establish explicit information propagation among representation tokens. To address this, we further propose a \emph{Progressive Representation Composition (PRC)} mechanism, which enables semantic refinement by progressively accumulating contextual knowledge layer by layer.

At each Transformer layer $i \in \{J+1, \dots, L\}$, the representation tokens for modality $m \in \{v, t\}$ are composed from two sources: the newly introduced tokens $R_{i-1}^m$, and the output tokens from the previous layer $O_{i-1}^m$. 

Specifically, for the image encoder:
\begin{align}
[c_i, E_i] &= \mathcal{V}_i([c_{i-1}, E_{i-1}]) \nonumber \\
&\hspace{2.4cm} i = 1, \cdots, J{-}1 \\
[c_i, O_i^v, E_i] &= \mathcal{V}_i([c_{i-1}, R_{i-1}^v, E_{i-1}]) \quad i = J \\
[c_i, O_i^v, E_i] &= \mathcal{V}_i([c_{i-1}, \beta R_{i-1}^v + (1-\beta)O_{i-1}^v, \nonumber \\
&\hspace{1cm} E_{i-1}]) \quad i = J+1, \cdots, L
\end{align}
where $\beta$ is a transfer coefficient. The final $O_L^v$ corresponds to the $R_L^v$ in \cref{R_L^v}, which is used to extract the features of the representation tokens. In the visual modality, during the interaction of representation tokens with other tokens at each layer, they capture instance-level knowledge. The PRC mechanism plays a crucial role in passing this valuable instance-level knowledge to subsequent layers. By doing so, it enriches the model's understanding of individual instances, which in turn significantly enhances the model's generalization ability across various tasks.

For the text encoder:
\begin{align}
[b_i, T_i, e_i] &= \mathcal{W}_i([b_{i-1}, T_{i-1}, e_{i-1}]) \nonumber \\
&\hspace{2cm} i = 1, \cdots, J{-}1 \\
[b_i, O_i^t, T_i, e_i] &= \mathcal{W}_i([b_{i-1}, R_{i-1}^t, T_{i-1}, e_{i-1}]) \nonumber \\
&\hspace{2cm} i = J \\
[b_i, O_i^t, T_i, e_i] &= \mathcal{W}_i([b_{i-1}, \beta R_{i-1}^t + (1-\beta)O_{i-1}^t, \nonumber \\
&\hspace{0cm} T_{i-1}, e_{i-1}]) \quad i = J+1, \cdots, L
\end{align}
The final $O_L^t$ corresponds to the $R_L^t$ in \cref{R_L^t}.

This additive composition allows the model to gradually refine its representations by integrating knowledge from preceding layers, facilitating instance-specific and domain-aware feature learning. Compared to isolated token optimization, PRC introduces explicit inter-layer token interactions and promotes stable and efficient training. As such, it serves as a natural complement to SRRA, further enhancing generalization and robustness.

\section{Experiments}
\label{sec:experiments}

\subsection{Tasks and Datasets}
We conduct four core evaluations to comprehensively assess our models' performance: base-to-novel generalization, cross-dataset evaluation, domain generalization, and few-shot learning. Except for few-shot learning, all experiments utilize a 16-shot setting, \textit{i.e.}, only 16 training examples per category.

\noindent \textbf{Base-to-Novel Generalization:} 
In this task, dataset classes are equally divided into base and novel classes. The model is trained exclusively on base classes and tested on both base and novel classes, allowing us to examine its transfer learning effectiveness on base classes as well as its ability to retain the inherent generalization or zero-shot capabilities of pre-trained VLMs for novel classes. We conduct this evaluation across 11 diverse image classification datasets: ImageNet \cite{imagenet}, Caltech101 \cite{caltech101}, OxfordPets \cite{oxford_pets}, StanfordCars \cite{stanford_cars}, Flowers102 \cite{flowers102}, Food101 \cite{food101}, FGVCAircraft \cite{fgvc_aircraft}, SUN397 \cite{sun397}, UCF101 \cite{ucf101}, DTD \cite{dtd}, and EuroSAT \cite{eurosat}. Details of this datasets are shown in \cref{datasets}.

\begin{table*}[t]
\centering
\caption{Comparison of MMRL and MMRL++ with previous state-of-the-art methods on base-to-novel generalization across 11 datasets. Bold values indicate the best results. MMRL and MMRL+ consistently enhance base class performance without compromising generalization.}
\label{base_to_novel}
\renewcommand\arraystretch{1.07}
\setlength{\tabcolsep}{2.5mm}{
\resizebox{0.95\textwidth}{!}{
    \begin{tabular}{@{}r|ccc|ccc|ccc|ccc@{}}
    \toprule
    \multirow{2}{*}{Method} &
      \multicolumn{3}{c|}{Average} &
      \multicolumn{3}{c|}{ImageNet} &
      \multicolumn{3}{c|}{Caltech101} &
      \multicolumn{3}{c}{OxfordPets} \\
               & Base  & Novel & HM    & Base           & Novel & HM    & Base           & Novel          & HM             & Base           & Novel          & HM             \\ \midrule
    CLIP       & 69.34 & 74.22 & 71.70 & 72.43          & 68.14 & 70.22 & 96.84          & 94.00          & 95.40          & 91.17          & 97.26          & 94.12          \\
    CoOp       & 82.69 & 63.22 & 71.66 & 76.47          & 67.88 & 71.92 & 98.00          & 89.81          & 93.73          & 93.67          & 95.29          & 94.47          \\
    CoOpOp     & 80.47 & 71.69 & 75.83 & 75.98          & 70.43 & 73.10 & 97.96          & 93.81          & 95.84          & 95.20          & 97.69          & 96.43          \\
    ProDA      & 81.56 & 72.30 & 76.65 & 75.40          & 70.23 & 72.72 & 98.27          & 93.23          & 95.68          & 95.43          & 97.83          & 96.62          \\
    KgCoOp     & 80.73 & 73.60 & 77.00 & 75.83          & 69.96 & 72.78 & 97.72          & 94.39          & 96.03          & 94.65          & 97.76          & 96.18          \\
    MaPLe      & 82.28 & 75.14 & 78.55 & 76.66          & 70.54 & 73.47 & 97.74          & 94.36          & 96.02          & 95.43          & 97.76          & 96.58          \\
    PromptSRC  & 84.26 & 76.10 & 79.97 & 77.60          & 70.73 & 74.01 & 98.10          & 94.03          & 96.02          & 95.33          & 97.30          & 96.30          \\
    ProVP      & 85.20 & 73.22 & 78.76 & 75.82          & 69.21 & 72.36 & 98.92          & 94.21          & 96.51          & 95.87          & 97.65          & \textbf{96.75} \\
    MetaPrompt & 83.65 & 75.48 & 79.09 & 77.52          & 70.83 & 74.02 & 98.13          & 94.58          & 96.32          & 95.53          & 97.00          & 96.26          \\
    TCP        & 84.13 & 75.36 & 79.51 & 77.27          & 69.87 & 73.38 & 98.23          & \textbf{94.67} & 96.42          & 94.67          & 97.20          & 95.92          \\
    MMA        & 83.20 & 76.80 & 79.87 & 77.31          & 71.00 & 74.02 & 98.40          & 94.00          & 96.15          & 95.40          & \textbf{98.07} & 96.72          \\ \midrule
    MMRL &
      \textbf{85.68} &
      77.16 &
      81.20 &
      \textbf{77.90} &
      71.30 &
      \textbf{74.45} &
      98.97 &
      94.50 &
      96.68 &
      \textbf{95.90} &
      97.60 &
      96.74 \\
    MMRL++ &
      85.53 &
      \textbf{78.32} &
      \textbf{81.77} &
      77.63 &
      \textbf{71.50} &
      74.44 &
      \textbf{99.07} &
      94.53 &
      \textbf{96.75} &
      95.60 &
      97.43 &
      96.51 \\ \midrule \midrule
    \multirow{2}{*}{Method} &
      \multicolumn{3}{c|}{StanfordCars} &
      \multicolumn{3}{c|}{Flowers102} &
      \multicolumn{3}{c|}{Food101} &
      \multicolumn{3}{c}{FGVCAircraft} \\
               & Base  & Novel & HM    & Base           & Novel & HM    & Base           & Novel          & HM             & Base           & Novel          & HM             \\ \midrule
    CLIP       & 63.37 & 74.89 & 68.65 & 72.08          & 77.80 & 74.83 & 90.10          & 91.22          & 90.66          & 27.19          & 36.29          & 31.09          \\
    CoOp       & 78.12 & 60.40 & 68.13 & 97.60          & 59.67 & 74.06 & 88.33          & 82.26          & 85.19          & 40.44          & 22.30          & 28.75          \\
    CoOpOp     & 70.49 & 73.59 & 72.01 & 94.87          & 71.75 & 81.71 & 90.70          & 91.29          & 90.99          & 33.41          & 23.71          & 27.74          \\
    ProDA      & 74.70 & 71.20 & 72.91 & 97.70          & 68.68 & 80.66 & 90.30          & 88.57          & 89.43          & 36.90          & 34.13          & 35.46          \\
    KgCoOp     & 71.76 & 75.04 & 73.36 & 95.00          & 74.73 & 83.65 & 90.50          & 91.70          & 91.09          & 36.21          & 33.55          & 34.83          \\
    MaPLe      & 72.94 & 74.00 & 73.47 & 95.92          & 72.46 & 82.56 & 90.71          & \textbf{92.05} & \textbf{91.38} & 37.44          & 35.61          & 36.50          \\
    PromptSRC  & 78.27 & 74.97 & 76.58 & 98.07          & 76.50 & 85.95 & 90.67          & 91.53          & 91.10          & 42.73          & 37.87          & 40.15          \\
    ProVP      & 80.43 & 67.96 & 73.67 & 98.42          & 72.06 & 83.20 & 90.32          & 90.91          & 90.61          & \textbf{47.08} & 29.87          & 36.55          \\
    MetaPrompt & 76.34 & 75.01 & 75.48 & 97.66          & 74.49 & 84.52 & \textbf{90.74} & 91.85          & 91.29          & 40.14          & 36.51          & 38.24          \\
    TCP        & 80.80 & 74.13 & 77.32 & 97.73          & 75.57 & 85.23 & 90.57          & 91.37          & 90.97          & 41.97          & 34.43          & 37.83          \\
    MMA        & 78.50 & 73.10 & 75.70 & 97.77          & 75.93 & 85.48 & 90.13          & 91.30          & 90.71          & 40.57          & 36.33          & 38.33          \\ \midrule
    MMRL       & 81.30 & 75.07 & 78.06 & \textbf{98.97} & 77.27 & 86.78 & 90.57          & 91.50          & 91.03          & 46.30          & 37.03          & 41.15          \\
    MMRL++ &
      \textbf{81.33} &
      \textbf{75.27} &
      \textbf{78.18} &
      98.53 &
      \textbf{77.90} &
      \textbf{87.01} &
      90.47 &
      91.73 &
      91.10 &
      46.40 &
      \textbf{38.77} &
      \textbf{42.24} \\ \midrule \midrule
    \multirow{2}{*}{Method} &
      \multicolumn{3}{c|}{SUN397} &
      \multicolumn{3}{c|}{DTD} &
      \multicolumn{3}{c|}{EuroSAT} &
      \multicolumn{3}{c}{UCF101} \\
               & Base  & Novel & HM    & Base           & Novel & HM    & Base           & Novel          & HM             & Base           & Novel          & HM             \\ \midrule
    CLIP       & 69.36 & 75.35 & 72.23 & 53.24          & 59.90 & 56.37 & 56.48          & 64.05          & 60.03          & 70.53          & 77.50          & 73.85          \\
    CoOp       & 80.60 & 65.89 & 72.51 & 79.44          & 41.18 & 54.24 & 92.19          & 54.74          & 68.69          & 84.69          & 56.05          & 67.46          \\
    CoOpOp     & 79.74 & 76.86 & 78.27 & 77.01          & 56.00 & 64.85 & 87.49          & 60.04          & 71.21          & 82.33          & 73.45          & 77.64          \\
    ProDA      & 78.67 & 76.93 & 77.79 & 80.67          & 56.48 & 66.44 & 83.90          & 66.00          & 73.88          & 85.23          & 71.97          & 78.04          \\
    KgCoOp     & 80.29 & 76.53 & 78.36 & 77.55          & 54.99 & 64.35 & 85.64          & 64.34          & 73.48          & 82.89          & 76.67          & 79.65          \\
    MaPLe      & 80.82 & 78.70 & 79.75 & 80.36          & 59.18 & 68.16 & 94.07          & 73.23          & 82.35          & 83.00          & 78.66          & 80.77          \\
    PromptSRC  & 82.67 & 78.47 & 80.52 & 83.37          & 62.97 & 71.75 & 92.90          & 73.90          & 82.32          & 87.10          & 78.80          & 82.74          \\
    ProVP      & 80.67 & 76.11 & 78.32 & 83.95          & 59.06 & 69.34 & \textbf{97.12} & 72.91          & 83.29          & \textbf{88.56} & 75.55          & 81.54          \\
    MetaPrompt & 82.26 & 79.04 & 80.62 & 83.10          & 58.05 & 68.35 & 93.53          & 75.21          & 83.38          & 85.33          & 77.72          & 81.35          \\
    TCP        & 82.63 & 78.20 & 80.35 & 82.77          & 58.07 & 68.25 & 91.63          & 74.73          & 82.32          & 87.13          & \textbf{80.77} & 83.83          \\
    MMA        & 82.27 & 78.57 & 80.38 & 83.20          & 65.63 & 73.38 & 85.46          & 82.34          & 83.87          & 86.23          & 80.03          & 82.20          \\ \midrule
    MMRL &
      \textbf{83.20} &
      79.30 &
      81.20 &
      \textbf{85.67} &
      65.00 &
      73.82 &
      95.60 &
      80.17 &
      87.21 &
      88.10 &
      80.07 &
      \textbf{83.89} \\
    MMRL++ &
      83.03 &
      \textbf{79.60} &
      \textbf{81.28} &
      85.47 &
      \textbf{65.97} &
      \textbf{74.46} &
      95.93 &
      \textbf{88.27} &
      \textbf{91.94} &
      87.37 &
      80.53 &
      83.81 \\ \bottomrule
    \end{tabular}
    }
}
\end{table*}

\noindent \textbf{Cross-Dataset Evaluation:} This evaluation measures the model’s transferability to new, unseen datasets. Following CoCoOp \cite{cocoop}, we train the model on all 1000 ImageNet classes in a few-shot setting and then directly apply it, without further fine-tuning, to other datasets to assess its cross-dataset generalization. We employ the same datasets as in the base-to-novel generalization task.

\noindent \textbf{Domain Generalization:} In this setting, we assess the resilience of the ImageNet-trained model to domain shifts and its generalization to out-of-distribution data. Specifically, we use ImageNet as the training dataset and evaluate on four variants—ImageNetV2 \cite{imagenetv2}, ImageNet-Sketch \cite{imagenet_sketch}, ImageNet-A \cite{imagenet_a}, and ImageNet-R \cite{imagenet_r}—each introducing different types of domain variation. Details of this datasets are shown in \cref{datasets}.

\noindent \textbf{Few-Shot Learning:} This evaluation examines the model's transfer learning capability in limited-data scenarios, independent of its generalization performance. The model is trained on subsets of the training data with 1, 2, 4, 8, and 16 examples (shots) per class and subsequently evaluated on the full test sets.

\subsection{Implementation Details}
We follow prior studies~\cite{coop, cocoop, prograd, kgcoop, maple, tcp, mma} and adopt a 16-shot learning setting for all experiments unless specified otherwise (e.g., few-shot learning tasks). The ViT-B/16~\cite{vit} variant of CLIP serves as the visual backbone across all setups. Hand-crafted text prompts from previous methods~\cite{clip, coop, tip-adapter} are employed, with details provided in \cref{datasets}. Models are optimized using AdamW with an initial learning rate of 0.001 and trained with mixed precision to accelerate computation. For the larger ImageNet dataset, we use a batch size of 32, while a batch size of 4 is applied to all other datasets. Training spans 5 epochs on ImageNet for base-to-novel generalization, and 10 epochs on the remaining datasets. For cross-dataset evaluation and domain generalization, we train for a single epoch on ImageNet. In few-shot learning, training is conducted over 5 epochs on ImageNet and 100 epochs for the other datasets. All results are averaged over three independent runs, with experiments performed on a single NVIDIA RTX 4090 GPU.

The representation space is initialized from a zero-mean Gaussian distribution with a standard deviation of 0.02 and a dimension $d_r=512$. We introduce $K=5$ representation tokens beginning at the 6-th Transformer layer ($J=6$). The low-rank dimensions $r_1$ and $r_2$ are set to 4 and 64, respectively. The balance weight $\alpha$ is fixed at 0.7, with the configurations for $\lambda$ and $\beta$ provided in the \textbf{Supplementary Material}.

\subsection{Base-to-Novel Generalization}

In this experiment, we compare MMRL and MMRL++ against several leading models, including the zero-shot CLIP baseline and state-of-the-art prompt learning methods: CoOp \cite{coop}, CoCoOp \cite{cocoop}, ProDA \cite{proda}, KgCoOp \cite{kgcoop}, MaPLe \cite{maple}, PromptSRC \cite{promptsrc}, ProVP \cite{provp}, MetaPrompt \cite{metaprompt}, TCP \cite{tcp}, and the multimodal adapter-based method MMA \cite{mma}. For fairness, we exclude methods that leverage large language models for strong prompt priors \cite{coprompt, hpt, argue, lweib} or distillation approaches \cite{promptkd} that utilize the full unlabeled dataset.

\cref{base_to_novel} provides detailed results for \textbf{Base} and \textbf{Novel} classes across 11 datasets, along with the balanced harmonic mean (\textbf{HM}) of their accuracies. Key findings include:

\noindent \textbf{New SOTA Performance.} MMRL and MMRL++ achieve superior average performance across all datasets. Specifically, MMRL yields improvements of 2.48\%, 0.36\%, and 1.33\% on Base, Novel, and HM metrics respectively, over the previous best model MMA. MMRL++ further enhances generalization, achieving gains of 2.33\% (Base), 1.52\% (Novel), and 1.90\% (HM), establishing a new state-of-the-art across all three metrics.

\noindent \textbf{Enhanced Transfer Learning with Strong Generalizability.} Notably, both MMRL and MMRL++ enhance generalizability while significantly boosting base accuracy, effectively improving transfer learning capabilities across downstream datasets. Although MMRL and MMRL++ do not simultaneously achieve the highest accuracy on Base or Novel on some datasets, they always far outperform the other models on one of the metrics to achieve the highest HM. For instance, on EuroSAT, MMRL underperforms MMA by 2.17\% in the Novel classes but outperforms it by 10.14\% in the Base classes. On the other hand, on FGVCAircraft, although MMRL++’s Base accuracy trails PromptVP by 0.68\%, yet it achieves a significant 8.90\% gain in Novel accuracy!

\noindent \textbf{Enhanced Generalization via SRRA and PRC.} While MMRL++ slightly underperforms MMRL in Base accuracy (by 0.15\%, which may be attributed to MMRL++ employing fewer trainable parameters), it achieves a notable improvement of 1.16\% in Novel accuracy. This indicates that the proposed SRRA and PRC modules significantly contribute to better generalization without severely compromising Base performance.

\begin{table*}[t]
\centering
\caption{Comparison of MMRL and MMRL++ with previous state-of-the-art methods on cross-dataset evaluation across 10 datasets.}
\label{cross_dataset}
\resizebox{0.98\textwidth}{!}{
    \begin{tabular}{@{}rc|ccccccccccc@{}}
    \toprule
     &
      Source &
      \multicolumn{11}{c}{Target} \\ \cmidrule(l){2-13} 
     &
      \rotatebox{60}{ImageNet} &
      \rotatebox{60}{Average} &
      \rotatebox{60}{Caltech101} &
      \rotatebox{60}{OxfordPets} &
      \rotatebox{60}{StanfordCars} &
      \rotatebox{60}{Flowers101} &
      \rotatebox{60}{Food101} &
      \rotatebox{60}{FGVCAircraft} &
      \rotatebox{60}{SUN397} &
      \rotatebox{60}{DTD} &
      \rotatebox{60}{EuroSAT} &
      \rotatebox{60}{UCF101} \\ \cmidrule(l){2-13} 
    CoOp &
      71.51 &
      63.88 &
      93.70 &
      89.14 &
      64.51 &
      68.71 &
      85.30 &
      18.47 &
      64.15 &
      41.92 &
      46.39 &
      66.55 \\
    CoOpOp &
      71.02 &
      65.74 &
      94.43 &
      90.14 &
      65.32 &
      71.88 &
      86.06 &
      22.94 &
      67.36 &
      45.73 &
      45.37 &
      68.21 \\
    MaPLe &
      70.72 &
      66.30 &
      93.53 &
      90.49 &
      65.57 &
      72.23 &
      86.20 &
      24.74 &
      67.01 &
      46.49 &
      48.06 &
      68.69 \\
    PromptSRC &
      71.27 &
      65.81 &
      93.60 &
      90.25 &
      65.70 &
      70.25 &
      86.15 &
      23.90 &
      67.10 &
      \textbf{46.87} &
      45.50 &
      68.75 \\
    TCP &
      71.40 &
      66.29 &
      93.97 &
      91.25 &
      64.69 &
      71.21 &
      86.69 &
      23.45 &
      67.15 &
      44.35 &
      51.45 &
      68.73 \\
    MMA &
      71.00 &
      66.61 &
      93.80 &
      90.30 &
      66.13 &
      72.07 &
      86.12 &
      25.33 &
      \textbf{68.17} &
      46.57 &
      49.24 &
      68.32 \\ \midrule
    MMRL &
      \textbf{72.03} &
      67.25 &
      \textbf{94.67} &
      \textbf{91.43} &
      66.10 &
      72.77 &
      86.40 &
      \textbf{26.30} &
      67.57 &
      45.90 &
      \textbf{53.10} &
      68.27 \\
    MMRL++ &
      71.87 &
      \textbf{67.49} &
      94.63 &
      \textbf{91.43} &
      \textbf{66.60} &
      \textbf{73.53} &
      \textbf{86.73} &
      26.07 &
      67.77 &
      46.13 &
      53.00 &
      \textbf{69.03} \\ \bottomrule
    \end{tabular}
    }
\end{table*}

\begin{table}[t]
\centering
\caption{Comparison of MMRL and MMRL++ with previous state-of-the-art methods on domain generalization across 4 datasets.}
\label{domain_generalization}
    \footnotesize
    \begin{tabular}{@{}rc|cccc@{}}
    \toprule
              & Source         & \multicolumn{4}{c}{Target}                                        \\ \cmidrule(l){2-6} 
              & ImNet          & -V2            & -S             & -A             & -R             \\ \cmidrule(l){2-6} 
    CLIP      & 66.73          & 60.83          & 46.15          & 47.77          & 73.96          \\
    CoOp      & 71.51          & 64.20          & 47.99          & 49.71          & 75.21          \\
    CoOpOp    & 71.02          & 64.07          & 48.75          & 50.63          & 76.18          \\
    MaPLe     & 70.72          & 64.07          & 49.15          & 50.90          & 76.98          \\
    PromptSRC & 71.27          & 64.35          & \textbf{49.55} & 50.90          & \textbf{77.80} \\
    MMA       & 71.00          & 64.33          & 49.13          & 51.12          & 77.32          \\ \midrule
    MMRL      & \textbf{72.03} & 64.47          & 49.17          & \textbf{51.20} & 77.53          \\
    MMRL++    & 71.87          & \textbf{64.67} & 49.30          & 51.00          & 77.43          \\ \bottomrule
    \end{tabular}
\end{table}

\subsection{Cross-Dataset Evaluation}
As shown in \cref{cross_dataset}, MMRL and MMRL++ exhibit strong generalization capabilities across a wide range of target datasets. MMRL achieves an average accuracy of 67.25\%, outperforming previous state-of-the-art methods such as MMA (66.61\%) and TCP (66.29\%). Notably, MMRL improves performance on ImageNet by 1.03\% over MMA, indicating not only enhanced robustness on the target domain but also improved transferability from the source domain. Building on this, MMRL++ further strengthens cross-dataset generalization, achieving the highest average accuracy of 67.49\%. These results underscore the effectiveness of our proposed framework in striking a strong balance between task-specific adaptation and generalization across domains.

\begin{table}[t]
\centering
\caption{Performance with different model variants across 11 datasets.}
\label{ablation_variants}
    \begin{tabular}{c|ccc}
    \toprule
    Variants                & Base           & Novel          & HM             \\ \midrule
    w/o L-Branch            & 79.78          & 74.97          & 77.30          \\
    w/o V-Branch            & 81.87          & 75.00          & 78.28          \\
    w/o DS-Base             & 83.10          & \textbf{78.32} & 80.64          \\
    w/o DS-Novel            & 85.53          & 76.93          & 81.00          \\
    w/o RS                  & \textbf{85.57} & 76.94          & 81.02          \\
    $\text{MMRL}^\dagger++$ & 85.53          & 76.76          & 80.91          \\ \midrule
    \rowcolor[HTML]{EFEFEF} 
    MMRL++                  & 85.53          & \textbf{78.32} & \textbf{81.77} \\ \bottomrule
    \end{tabular}
\end{table}

\begin{figure*}[t]
\centering
  \includegraphics[width=1.0\linewidth]{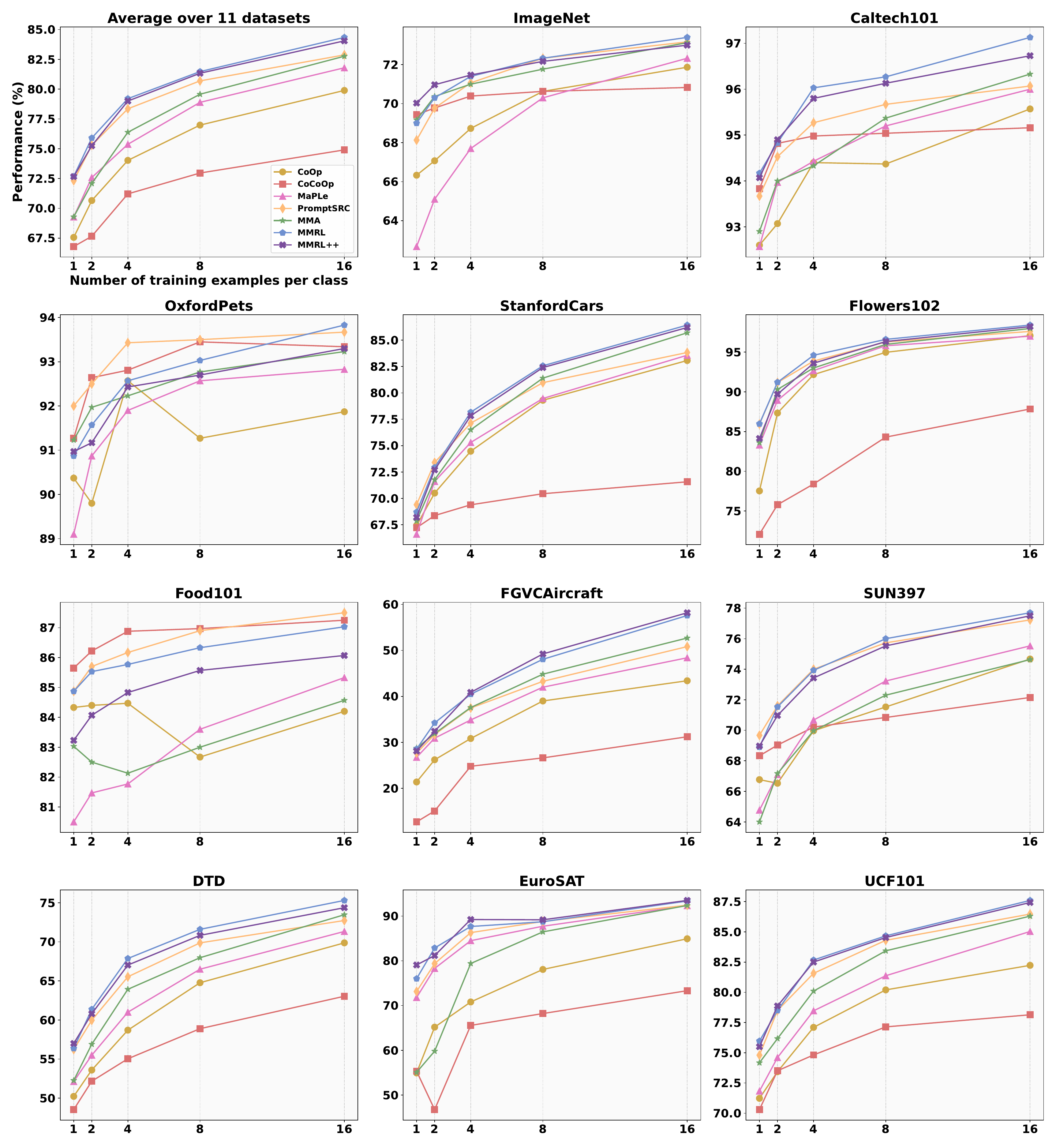}
  \caption{Comparison of MMRL and MMRL++ with previous state-of-the-art methods on few-shot learning across 11 datasets. Detailed results on all 11 datasets are provided in the \textbf{Supplementary Material}.}
  \label{few_shot_figure}
\end{figure*}

\begin{table*}[t]
\centering
\caption{Impact of the low-rank space dimension $r_1$ and $r_2$, averaged across 11 datasets, and the effects of $\lambda$ and $\beta$ evaluated on ImageNet.}
\label{ablation_rank_lambda_beta}
    \begin{subtable}{0.49\linewidth}
    \centering
    \resizebox{1\textwidth}{!}{
        \begin{tabular}{c|ccc||c|ccc}
        \toprule
        $r_1$ & Base  & Novel & HM    & $r_2$ & Base  & Novel & HM    \\ \midrule
        1     & 85.42 & 77.68 & 81.36 & 4     & 84.16 & 76.57 & 80.19 \\
        \cellcolor[HTML]{EFEFEF}4 &
          \cellcolor[HTML]{EFEFEF}85.53 &
          \cellcolor[HTML]{EFEFEF}\textbf{78.32} &
          \cellcolor[HTML]{EFEFEF}\textbf{81.77} &
          16 &
          84.74 &
          77.09 &
          80.73 \\
        64 &
          \textbf{85.62} &
          77.49 &
          81.35 &
          \cellcolor[HTML]{EFEFEF}64 &
          \cellcolor[HTML]{EFEFEF}\textbf{85.53} &
          \cellcolor[HTML]{EFEFEF}\textbf{78.32} &
          \cellcolor[HTML]{EFEFEF}\textbf{81.77} \\
        256   & 85.55 & 77.14 & 81.13 & 256   & 85.28 & 77.20 & 81.04 \\
        512   & 85.49 & 76.95 & 81.00 & 512   & 85.08 & 76.25 & 80.42 \\ \bottomrule
        \end{tabular} 
        }
    \end{subtable}
    \begin{subtable}{0.49\linewidth}
    \centering
    \resizebox{0.985\textwidth}{!}{
        \begin{tabular}{c|ccc||c|ccc}
        \toprule
        $\lambda$ &
          Base &
          Novel &
          HM &
          $\beta$ &
          Base &
          Novel &
          HM \\ \midrule
        0.0 &
          77.50 &
          71.23 &
          74.23 &
          0.0 &
          77.23 &
          71.07 &
          74.02 \\
        \cellcolor[HTML]{EFEFEF}0.2 &
          \cellcolor[HTML]{EFEFEF}\textbf{77.63} &
          \cellcolor[HTML]{EFEFEF}\textbf{71.50} &
          \cellcolor[HTML]{EFEFEF}\textbf{74.44} &
          0.3 &
          77.40 &
          71.43 &
          74.30 \\
        0.5 &
          77.57 &
          71.40 &
          74.36 &
          0.6 &
          77.60 &
          71.37 &
          74.35 \\
        2.0 &
          77.37 &
          71.00 &
          74.05 &
          \cellcolor[HTML]{EFEFEF}0.9 &
          \cellcolor[HTML]{EFEFEF}\textbf{77.63} &
          \cellcolor[HTML]{EFEFEF}\textbf{71.50} &
          \cellcolor[HTML]{EFEFEF}\textbf{74.44} \\
        4.0 &
          77.17 &
          70.63 &
          73.76 &
          1.0 &
          77.53 &
          71.33 &
          74.30 \\ \bottomrule
        \end{tabular}
        }
    \end{subtable}
\end{table*}

\begin{figure*}[t]
\centering
  \includegraphics[width=1.0\linewidth]{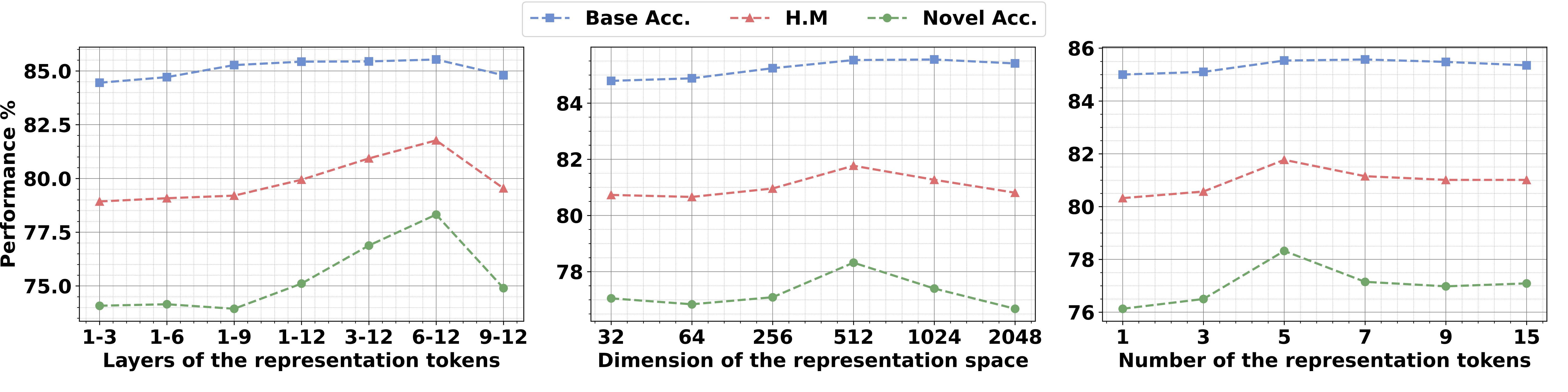}
    \caption{Performance across different insertion layers for representation tokens, along with the effects of the representation space dimension $d_r$ and the number of representation tokens $K$.}
  \label{ablation_layer_dimension_K}
\end{figure*}

\begin{table}[]
\centering
\caption{Ablation study on the SRRA and PRC modules in MMRL++, compared to the MMRL baseline.}
\label{ablation_srra_prc}
\setlength{\tabcolsep}{5.5pt}{
    \begin{tabular}{cc|ccc|c}
    \toprule
    PRC          & SRRA         & Base           & Novel & HM    & \begin{tabular}[c]{@{}c@{}}Trainable\\ Params\end{tabular} \\ \midrule
    $\times$     & $\times$     & 85.68          & 77.16 & 81.20 & 4.992M                                                     \\
    $\checkmark$ & $\times$     & \textbf{85.79} & 77.42 & 81.39 & 4.992M                                                     \\
    $\times$     & $\checkmark$ & 85.46          & 77.48 & 81.27 & \textbf{0.813M}                                            \\
    \rowcolor[HTML]{EFEFEF} 
    $\checkmark$ & $\checkmark$ & \cellcolor[HTML]{EFEFEF}85.53 & \cellcolor[HTML]{EFEFEF}\textbf{78.32} & \cellcolor[HTML]{EFEFEF}\textbf{81.77} & \textbf{0.813M} \\ \bottomrule
    \end{tabular}
    }
\end{table}

\subsection{Domain Generalization}
As summarized in \cref{domain_generalization}, MMRL and MMRL++ attains top performance on 2 out of the 4 domain-shifted datasets, showcasing there robust generalization capability across diverse domains. 

\subsection{Few-Shot Learning}
As illustrated in \cref{few_shot_figure}, MMRL consistently achieves the highest average performance across 11 datasets under all shot settings, with its performance advantage becoming more pronounced as the number of shots increases. MMRL++ closely follows, ranking second across all settings. This trend highlights the strong transfer learning capabilities of both MMRL and MMRL++, even in data-scarce scenarios, confirming their robustness and effectiveness in few-shot scenarios.

\subsection{Ablation Analysis}
All ablation experiments, except for the analysis of $\lambda$ and $\beta$, are conducted under the base-to-novel generalization setting across 11 datasets, with averaged results reported. The analysis of $\lambda$ and $\beta$ is performed on ImageNet; please refer to the \textbf{Supplementary Material} for comprehensive results across all datasets.

\noindent \textbf{Variants of MMRL++:} As presented in \cref{ablation_variants}, the performance of MMRL++ variants highlights the contributions of its key components. In the `w/o L-Branch' and `w/o V-Branch' variants, where representation tokens are inserted into only a single modality encoder, a noticeable drop in performance is observed. This outcome underscores the importance of multimodal interaction and representation features in MMRL++. The `w/o DS-Base' variant, which omits the Decoupling Strategy and relies solely on class features for base class evaluation, exhibits substantial performance degradation, emphasizing the critical role of representation features in capturing downstream knowledge. In the `w/o DS-Novel' variant, which uses both feature types for novel class evaluation, performance also declines notably. This suggests that class tokens encode more generalized knowledge, while representation tokens primarily capture dataset-specific information with limited generalizability. The `w/o RS' variant, which removes the Representation Space and initializes textual and visual tokens independently, forgoes multimodal learning. Although this unimodal strategy improves base class performance, it significantly hampers generalization to novel classes, highlighting the necessity of multimodal learning for robust generalization. Lastly, $\text{MMRL}^\dagger++$ implements a biased multimodal learning scheme analogous to MaPLe \cite{maple}, where text-side tokens are randomly initialized and mapped to the visual side. The results show that this biased approach underperforms compared to MMRL++’s balanced multimodal learning framework.

\noindent \textbf{Effectiveness of SRRA and PRC Modules (MMRL++ vs. MMRL):} Table~\ref{ablation_srra_prc} presents an ablation study evaluating the individual and combined contributions of the SRRA and PRC modules within the MMRL++ framework. The baseline configuration (without SRRA or PRC, i.e., MMRL) achieves an HM score of 81.20 with 4.992M trainable parameters. Incorporating the PRC module alone enhances both base and novel class recognition, increasing the HM score to 81.39. In contrast, enabling only the SRRA module reduces the parameter count to 0.813M—an 84\% reduction—while maintaining competitive performance (HM = 81.27) and improving novel class accuracy, underscoring the parameter efficiency of SRRA. When both SRRA and PRC are enabled, the model not only remains lightweight but also achieves the highest overall performance (HM = 81.77), driven by the best novel class accuracy (78.32). These results confirm that SRRA substantially improves both parameter efficiency and generalization, and that the PRC module further enhances generalization when used in combination with SRRA.

\noindent \textbf{Layer for Representation Token Insertion:} As illustrated in \cref{ablation_layer_dimension_K} (left), model performance on novel classes degrades when representation tokens are inserted into lower encoder layers. This observation is consistent with the design rationale of MMRL++: higher encoder layers are more effective at capturing dataset-specific, discriminative features, whereas lower layers tend to preserve general-purpose representations. Furthermore, inserting representation tokens into lower layers also results in decreased performance on base classes, indicating that these layers are less responsive to task-specific adaptation. Performance improves as the tokens are inserted into progressively higher layers; however, inserting them too high in the hierarchy again leads to performance drops—likely due to fewer trainable parameters and diminished influence over CLIP’s critical layers.

\begin{table*}[h]
\centering
\renewcommand\arraystretch{1}
\caption{All methods are trained on a single NVIDIA RTX 4090 GPU using the ImageNet dataset. Each model is implemented using publicly available code and default configurations as described in their respective papers \cite{maple, promptsrc, provp, metaprompt, tcp, mma}. `V-L' denotes vision-language interaction, indicating that efficient fine-tuning incorporates interactions between visual and textual modalities before prediction. `V, L' signifies separate fine-tuning of each modality without inter-modal interaction before prediction, while `V' or `L' refers to fine-tuning limited to the visual or textual modality alone. `Train time' is reported as both time per image and the total duration for training the full dataset (16-shots), while `FPS (100 BS)' indicates frames per second with a batch size of 100 during inference.}
\label{computational_cost}
\resizebox{0.85\textwidth}{!}{
    \begin{tabular}{@{}r|ccccc|ccc@{}}
    \toprule
    \multirow{2}{*}{Method} & \multirow{2}{*}{Modality} & Params & Train time & Train time & FPS & \multirow{2}{*}{Base} & \multirow{2}{*}{Novel} & \multirow{2}{*}{HM} \\
               &     & (trainable) & (ms/image) & (minute/all) & (100 BS) &       &       &       \\ \midrule
    MaPLe      & V-L & 3.555M      & 39.5       & 26.4         & 1757.6   & 82.28 & 75.14 & 78.55 \\
    PromptSRC  & V,L & 0.046M      & 40.0       & 106.8        & 1764.2   & 84.26 & 76.10 & 79.97 \\
    ProVP      & V   & 0.147M      & 6.4        & 107.2        & 928.9    & 85.20 & 73.22 & 78.76 \\
    MetaPrompt & V,L & 0.031M      & 30.7       & 32.8         & 659.8    & 83.65 & 75.48 & 79.09 \\
    TCP        & L   & 0.332M      & 5.3        & 17.7         & 950.6    & 84.13 & 75.36 & 79.51 \\
    MMA        & V-L & 0.675M      & 2.2        & 1.5          & 688.5    & 83.20 & 76.80 & 79.87 \\ \midrule
    MMRL       & V-L & 4.992M      & 5.3        & 3.6          & 762.4    & 85.68 & 77.16 & 81.20 \\
    MMRL++     & V-L & 0.813M      & 5.4        & 3.7          & 761.3    & 85.53 & 78.32 & 81.77 \\
    MMRL*++    & V-L & 0.170M      & 5.4        & 3.7          & 766.7    & 84.79 & 77.05 & 80.73 \\ \bottomrule
    \end{tabular}
}
\end{table*}

\begin{table*}[t]
\centering
\caption{Training stability comparison between MMRL and MMRL++, measured by mean $\pm$ standard deviation (\%).}
\label{training_stability}
\resizebox{0.8\textwidth}{!}{
    \begin{tabular}{@{}r|ccc|ccc@{}}
    \toprule
    \multirow{2}{*}{Method} & \multicolumn{3}{c|}{ImageNet} & \multicolumn{3}{c}{EuroSAT} \\
                            & Base     & Novel     & HM     & Base     & Novel    & HM    \\ \midrule
    MMRL   & $77.90 \pm \textbf{0.08}$ & $71.30 \pm 0.28$          & 74.45 & $95.60 \pm 0.33$          & $80.17 \pm 5.05$          & 87.21 \\
    MMRL++ & $77.63 \pm 0.09$ & $71.50 \pm \textbf{0.08}$ & 74.44 & $95.93 \pm \textbf{0.29}$ & $88.27 \pm \textbf{0.54}$ & 91.94 \\ \bottomrule
    \end{tabular}
}
\end{table*}

\noindent \textbf{Dimension of Representation Space $d_r$ and Number of Representation Tokens $K$:} As shown in \cref{ablation_layer_dimension_K} (middle), performance initially improves and then declines as $d_r$ increases. This decline is likely due to overfitting caused by an excessively complex representation space. In \cref{ablation_layer_dimension_K} (right), increasing $K$ slightly improves accuracy on base classes, attributed to the added learnable parameters. For novel classes, accuracy initially rises with larger $K$ but eventually drops, indicating that an excessive number of tokens may induce overfitting and impair the model's generalization ability.
\begin{table}[t]
\centering
\small
\caption{Results for different $\alpha$, averaged across 11 datasets.}
\label{ablation_alpha}
    \begin{tabular}{c|ccc}
    \toprule
    $\alpha$ & Base           & Novel          & HM             \\ \midrule
    0.0      & 83.09          & 75.32          & 79.01          \\
    0.3      & 84.60          & 77.64          & 80.97          \\
    0.5      & 85.23          & 77.93          & 81.42          \\
    \rowcolor[HTML]{EFEFEF} 
    0.7      & \textbf{85.53} & \textbf{78.32} & \textbf{81.77} \\
    1.0      & 83.75          & 77.11          & 80.29          \\ \bottomrule
    \end{tabular}
\end{table}

\noindent \textbf{Dimension of Low-Rank Matrices $r_1$ and $r_2$:} As shown in \cref{ablation_rank_lambda_beta} (left), varying the dimensions of the low-rank matrices $r_1$ and $r_2$ significantly influences performance. For both $r_1$ and $r_2$, accuracy on base classes improves as the rank increases from 1 to 64. The configuration with $r_1 = 4$ and $r_2 = 64$ achieves the highest HM score of 81.77, indicating a well-balanced trade-off between base and novel class performance. However, increasing the ranks beyond 64 yields diminishing returns, often resulting in either marginal degradation on base class accuracy or more pronounced declines on novel class performance. This decline is likely due to overfitting and the introduction of redundant parameters that do not contribute to generalizable feature representations. Notably, using low-rank matrices with smaller dimensions (e.g., $r_1 = 4$) substantially reduces the number of trainable parameters with minimal impact on performance. This parameter efficiency lowers computational cost and may enhance generalization by mitigating over-parameterization. Thus, low-rank approximation offers an effective means to lightweight the model while preserving adaptability across tasks.

\noindent \textbf{Penalty Coefficient $\lambda$ and Transfer Coefficient $\beta$:} The penalty coefficient $\lambda$ controls the strength of regularization by aligning both the class token and text features with those of the frozen CLIP model. Increasing $\lambda$ enhances generalization but may limit the model's flexibility for task-specific adaptation. For ImageNet, $\lambda = 0.2$ offers the best trade-off, as shown in \cref{ablation_rank_lambda_beta} (right). The transfer coefficient $\beta$ regulates the flow of representation token information from earlier encoder layers. When $\beta = 0$, only information from the initial layer is propagated; when $\beta = 1$, representation tokens receive no information from preceding layers (i.e., the PRC module is disabled). Empirically, setting $\beta = 0.9$ yields optimal performance on ImageNet, as illustrated in \cref{ablation_rank_lambda_beta} (right).

\noindent \textbf{Balance Weight, $\alpha$:} The parameter $\alpha$ regulates the relative reliance on representation token features versus class token features. Lower values of $\alpha$ increase dependence on representation features, thereby raising the risk of overfitting due to the learnable projection layer. In contrast, higher values shift reliance toward class token features, which may reduce the model’s transferability. As shown in \cref{ablation_alpha}, the optimal value of $\alpha$ is 0.7.

\begin{figure*}[t]
    \centering
    \begin{subfigure}[b]{0.49\textwidth}
        \centering
        \includegraphics[width=\textwidth]{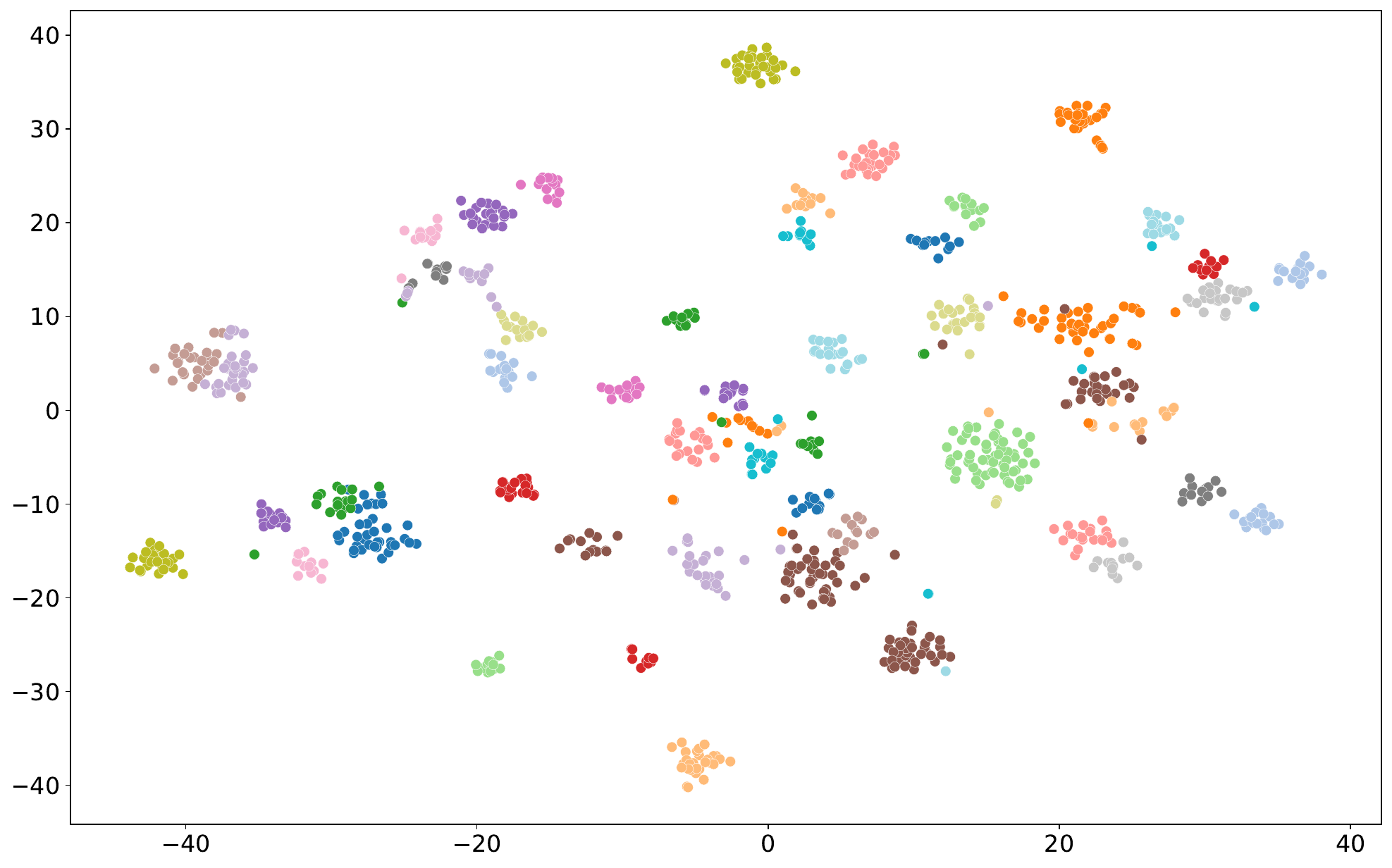}
        \caption{Class features on base classes}
        \label{tsne_cls_features_base}
    \end{subfigure}
    \begin{subfigure}[b]{0.49\textwidth}
        \centering
        \includegraphics[width=\textwidth]{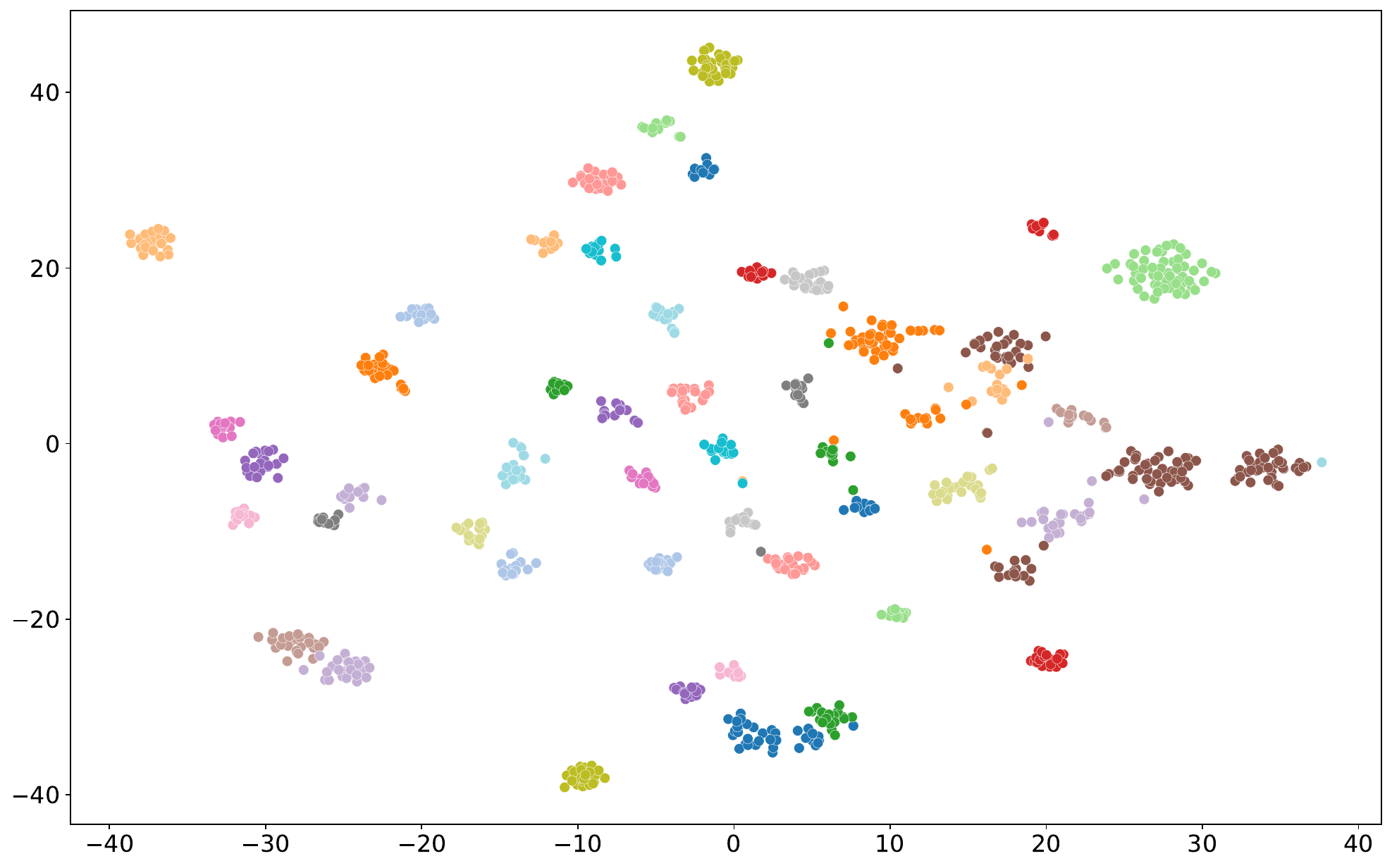}
        \caption{Representation features on base classes}
        \label{tsne_rep_features_base}
    \end{subfigure}   
    
    \begin{subfigure}[b]{0.49\textwidth}
        \centering
        \includegraphics[width=\textwidth]{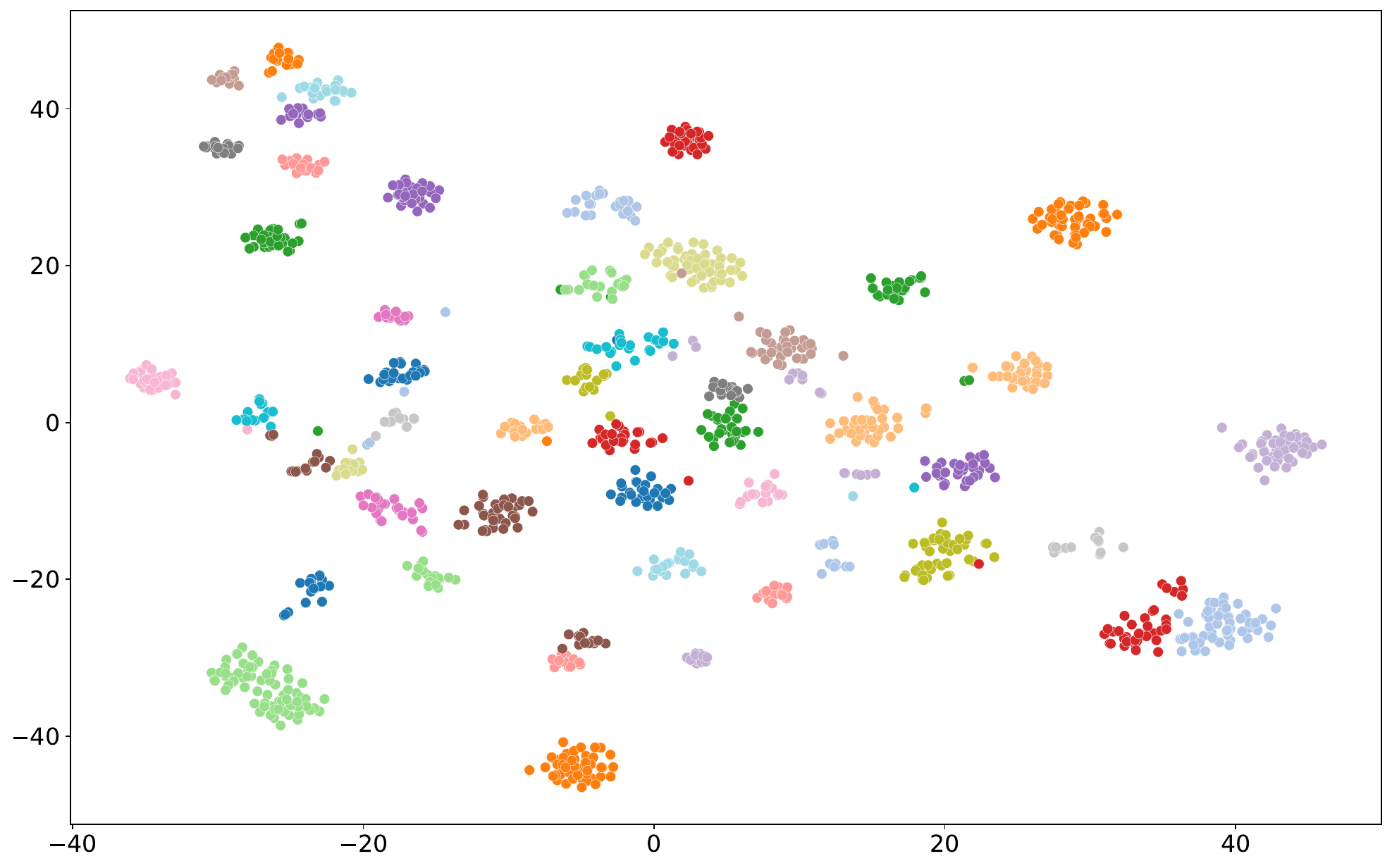}
        \caption{Class features on novel classes}
        \label{tsne_cls_features_new}
    \end{subfigure}
    \begin{subfigure}[b]{0.49\textwidth}
        \centering
        \includegraphics[width=\textwidth]{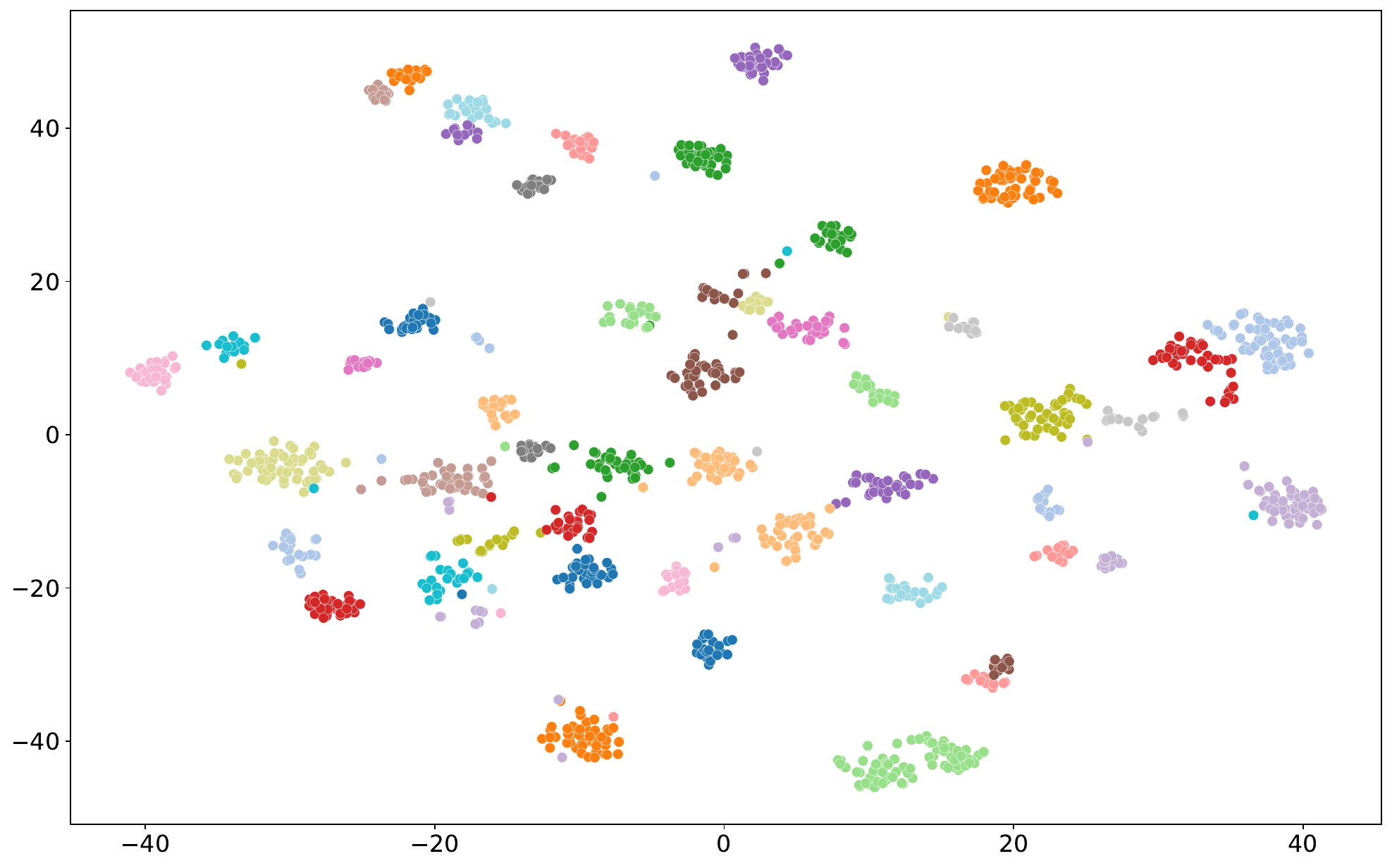}
        \caption{Representation features on novel classes}
        \label{tsne_rep_features_new}
    \end{subfigure}
    \caption{t-SNE visualizations of learned features from class token and representation tokens on the base and novel classes of the Flowers102 dataset.}
    \label{t-SNE}
\end{figure*}

\subsection{Further Analysis}

\subsubsection{Computational Cost}
Table \ref{computational_cost} summarizes the number of learnable parameters, training time per image, total training duration, inference speed (measured in frames per second, FPS, with a batch size of 100), and the final Base, Novel, HM metric for each approach. Our proposed models, MMRL and MMRL++, demonstrate a strong balance between computational efficiency and performance. Key observations include:
\begin{itemize}
    \item Models incorporating multimodal interaction mechanisms (\textit{e.g.}, MaPLe, MMA, MMRL, and MMRL++) generally have a higher parameter count than those without such mechanisms.
    \item MMRL, MMRL++, and the prior MMA approach exhibit significantly faster training speeds, reducing overall computational cost. While MaPLe and PromptSRC achieve higher inference speeds, their training durations are relatively longer. Notably, MMRL and MMRL++ offer faster inference than both MMA and MetaPrompt. 
    \item Although MMRL++ has fewer parameters due to low-rank decomposition and shared modules, operations such as multi-branch fusion and progressive composition introduce additional runtime overhead. As a result, MMRL++ shows a slight increase in training time compared to MMRL.
    \item To evaluate MMRL++ under constrained computational resources, we reduced the dimensionality of the representation space from 512 to 32. In this setting, MMRL*++ utilizes approximately one-quarter the parameters of MMA while still substantially outperforming the previous state-of-the-art.
\end{itemize}

\subsubsection{Training Stability}
We assess the training stability of MMRL and MMRL++ by reporting standard deviations across three random seeds on both ImageNet and EuroSAT datasets, as shown in \cref{training_stability}. Compared to MMRL, MMRL++ consistently achieves lower standard deviations, particularly on novel classes. Notably, on EuroSAT, where the limited number of classes increases the risk of instability under few-shot settings, the standard deviation for novel classes drops significantly from $5.05$ to $0.54$. These results demonstrate that the proposed SRRA and PRC mechanisms in MMRL++ effectively enhance training stability, leading to more reliable and consistent adaptation performance across diverse seeds.

\subsection{Feature Visualization}

To better understand our model’s behavior, we perform t-SNE visualizations of the learned features from both class tokens and representation tokens on base and novel classes. As shown in \cref{t-SNE}, for base classes, features extracted from representation tokens form more compact and well-separated clusters compared to those from class tokens, indicating stronger discriminative power and better task-specific adaptation. For novel classes, although both types of features exhibit reasonably good clustering in the t-SNE plots, the classification performance of representation tokens degrades significantly when computing logits against the text embeddings (see \cref{ablation_variants}). This indicates a weaker alignment between the representation features and the text space of CLIP. These observations further validate our decoupling strategy: class tokens are used for inference on novel classes due to their superior generalization and alignment with the pre-trained zero-shot space, while representation tokens enhance task-specific performance on base classes.

Importantly, this does not imply that the visual features from representation tokens are without value on novel classes. On the contrary, in scenarios where only visual information is required and textual features are not involved, more effective fusion of complementary representational and class features may become a valuable direction for future exploration.

\section{Conclusion}
In this work, we propose \textbf{MMRL}, a framework designed to enhance the generalization of VLMs by introducing a shared representation space that bridges the image and text modalities, while decoupling representation and class tokens to mitigate overfitting. Building upon this, \textbf{MMRL++} further improves scalability and stability through a Shared-Residual Representation Aligner, enabling parameter-efficient adaptation and knowledge sharing, and a Progressive Representation Composition mechanism that facilitates instance-specific knowledge propagation. Extensive experiments demonstrate that MMRL and MMRL++ achieve state-of-the-art performance in few-shot adaptation while maintaining strong generalization, establishing a new benchmark for efficient VLM transfer learning.

\noindent\textbf{Acknowledgements} This work was supported by National Natural Science Foundation of China under Grant 62176062.

\noindent\textbf{Data Availability.}
The datasets used in our paper are publicly available at the following links:
\begin{itemize}
\item ImageNet: \url{https://image-net.org/index.php}
\item Caltech101: \url{https://data.caltech.edu/records/mzrjq-6wc02} \item OxfordPets: \url{https://www.robots.ox.ac.uk/~vgg/data/pets/} 
\item StanfordCars: \url{https://ai.stanford.edu/~jkrause/cars/car_dataset.html} 
\item Flowers102: \url{https://www.robots.ox.ac.uk/~vgg/data/flowers/102/index.html} 
\item Food101: \url{https://data.vision.ee.ethz.ch/cvl/datasets_extra/food-101/} 
\item FGVCAircraft: \url{https://www.robots.ox.ac.uk/~vgg/data/fgvc-aircraft} 
\item SUN397: \url{https://vision.princeton.edu/projects/2010/SUN/} 
\item UCF101: \url{https://www.crcv.ucf.edu/data/UCF101.php} 
\item DTD: \url{https://www.robots.ox.ac.uk/~vgg/data/dtd} 
\item EuroSAT: \url{https://github.com/phelber/eurosat} 
\item ImageNetV2: \url{https://github.com/modestyachts/ImageNetV2} 
\item ImageNet-Sketch: \url{https://github.com/HaohanWang/ImageNet-Sketch} 
\item ImageNet-A: \url{ https://github.com/hendrycks/natural-adv-examples}
\item ImageNet-R: \url{https://github.com/hendrycks/imagenet-r} \end{itemize}


\bibliography{sn-bibliography}

\clearpage

\twocolumn[
  \begin{center}
    {\LARGE \textbf{Supplementary Material}}
  \end{center}
  \vspace{1em}
]

\begin{appendices}

\section{Detailed Results on All 11 Datasets of Ablation Analysis on $\lambda$}
As shown in \cref{ablation_lambda}, increasing the value of $\lambda$ generally improves model performance. The best or near-best results are typically achieved when $\lambda$ is set between 2 and 7 across most datasets. This trend indicates that incorporating the proposed regularization term positively influences adaptation. Notably, as $\lambda$ increases further, its impact on performance within the same dataset diminishes, suggesting reduced sensitivity to its variation. This observation implies that the model becomes more robust and less dependent on precise tuning of $\lambda$ at higher values.

\begin{table*}[t]
\centering
\caption{Performance across different values of $\lambda$ on 11 datasets, evaluated using the harmonic mean (HM) metric.}
\label{ablation_lambda}
\resizebox{1.0\textwidth}{!}{
    \begin{tabular}{@{}c|ccccccccccc@{}}
    \toprule
    $\lambda$ & ImageNet & Caltech101     & OxfordPets & StanfordCars & Flowers102 & Food101 & FGVCAircraft & SUN397         & DTD   & EuroSAT & UCF101         \\ \midrule
    0.0  & 74.23          & 96.17 & 96.10          & 76.30          & 84.05          & 90.70          & 39.58          & 80.19 & 67.53          & 90.95          & 81.69 \\
    0.01 & 74.32          & 96.32 & \textbf{96.51} & 75.95          & 83.22          & 90.70          & 39.32          & 79.83 & 67.95          & \textbf{91.94} & 81.69 \\
    0.1  & 74.41          & 96.50 & 96.49          & 76.44          & 84.47          & 90.89          & 39.78          & 80.23 & 68.98          & 89.72          & 82.28 \\
    0.2  & \textbf{74.44} & 96.39 & 96.21          & 76.72          & 85.35          & 90.94          & 40.89          & 80.51 & 70.11          & 88.64          & 82.15 \\
    0.5  & 74.36          & 96.49 & 96.39          & 77.12          & 85.54          & 91.06          & 41.54          & 80.64 & 71.79          & 85.94          & 82.73 \\
    1.0  & 74.30          & 96.63 & 96.32          & 77.28          & 85.86          & \textbf{91.10} & 41.39          & 80.96 & 72.58          & 81.65          & 83.11 \\
    2.0  & 74.05          & 96.45 & 96.23          & 77.80          & 86.23          & 91.05          & \textbf{42.24} & 81.23 & 73.19          & 80.29          & 83.70 \\
    3.0      & 73.88    & \textbf{96.75} & 96.48      & 77.84        & 86.45      & 91.06   & 41.92        & \textbf{81.28} & 73.59 & 80.39   & \textbf{83.81} \\
    4.0  & 73.76          & 96.54 & 96.32          & 77.90          & 86.44          & 91.06          & 41.67          & 81.08 & 73.58          & 78.04          & 83.40 \\
    5.0  & 73.65          & 96.70 & 96.16          & 77.81          & 86.54          & 91.03          & 41.65          & 81.00 & 73.93          & 77.34          & 83.69 \\
    6.0  & 73.53          & 96.53 & 96.34          & \textbf{78.18} & 86.72          & 91.00          & 41.53          & 80.97 & 73.91          & 77.50          & 83.49 \\
    7.0  & 73.48          & 96.52 & 96.26          & 78.00          & \textbf{87.01} & 91.03          & 41.52          & 80.96 & \textbf{74.46} & 77.64          & 83.57 \\
    10.0 & 73.29          & 96.55 & 96.43          & 77.75          & 86.60          & 90.95          & 41.40          & 80.81 & 73.57          & 77.79          & 83.42 \\ \bottomrule
    \end{tabular}
}
\end{table*}

\begin{table*}[t]
\centering
\caption{Performance across different values of $\beta$ on 11 datasets, evaluated using the harmonic mean (HM) metric.}
\label{ablation_beta}
\resizebox{1.0\textwidth}{!}{
    \begin{tabular}{@{}c|ccccccccccc@{}}
    \toprule
    $\beta$ & ImageNet       & Caltech101 & OxfordPets     & StanfordCars   & Flowers102 & Food101 & FGVCAircraft   & SUN397 & DTD            & EuroSAT & UCF101         \\ \midrule
    0.0 & 74.02 & 96.44          & 96.42          & 77.65 & 86.79          & 90.99          & 41.00 & 80.81          & 73.19 & 89.34          & 82.71 \\
    0.1 & 74.20 & 96.52          & 96.44          & 77.49 & 86.41          & \textbf{91.10} & 41.35 & 81.06          & 73.26 & 90.12          & 83.19 \\
    0.2 & 74.26 & 96.59          & 96.13          & 77.53 & 86.33          & 91.00          & 41.69 & 81.16          & 73.34 & \textbf{91.94} & 82.71 \\
    0.3 & 74.30 & 96.61          & 96.10          & 77.65 & 86.73          & 91.00          & 41.21 & 81.03          & 72.61 & 90.68          & 82.87 \\
    0.4 & 74.38 & 96.45          & 96.08          & 77.48 & \textbf{87.01} & 90.95          & 40.85 & 81.01          & 73.38 & 87.17          & 82.86 \\
    0.5 & 74.37 & 96.44          & 96.22          & 77.64 & 86.55          & 90.83          & 41.00 & \textbf{81.28} & 73.63 & 88.07          & 83.05 \\
    0.6 & 74.35 & \textbf{96.75} & 96.29          & 77.68 & 86.73          & 90.87          & 41.04 & 80.99          & 73.91 & 88.13          & 83.58 \\
    0.7     & 74.41          & 96.52      & \textbf{96.51} & \textbf{78.18} & 86.55      & 90.79   & 41.47          & 81.06  & 74.10          & 80.19   & 82.91          \\
    0.8 & 74.36 & 96.54          & \textbf{96.51} & 77.98 & 86.57          & 90.70          & 41.76 & 81.05          & 74.40 & 84.38          & 83.62 \\
    0.9     & \textbf{74.44} & 96.55      & 96.29          & 78.08          & 86.38      & 90.76   & \textbf{42.24} & 81.08  & \textbf{74.46} & 78.94   & \textbf{83.81} \\
    1.0 & 74.30 & 96.51          & 96.07          & 78.05 & 86.74          & 90.72          & 42.00 & 81.07          & 73.44 & 82.67          & 83.56 \\ \bottomrule
    \end{tabular}
}
\end{table*}

\section{Detailed Results on All 11 Datasets of Ablation Analysis on $\beta$}
The coefficient $\beta$ in our Progressive Representation Composition (PRC) module governs the strength of information flow from prior layer of representation tokens to the next. Specifically, smaller values of $\beta$ correspond to stronger transfer. As shown in \cref{ablation_beta}, most datasets achieve optimal or near-optimal performance when $\beta$ lies in the range $[0.5, 0.9]$, suggesting that moderate transfer strength strikes the best balance between knowledge propagation and noise control. These results highlight the importance of appropriately tuning $\beta$ to ensure effective intra-modal interaction and progressive refinement of semantic representations across layers.

It is important to note that the results obtained by setting $\beta = 1.0$ in this experiment (i.e., enabling SRRA while disabling PRC) are not directly comparable to those reported in the ablation study in the main paper (Table 7), where only SRRA is enabled. This discrepancy arises because the current experiment is based on MMRL++, where setting $\beta = 0$ disables PRC—functionally equivalent to introducing the SRRA module without parameter tuning. In contrast, the results in the main text are based on integrating SRRA into MMRL, with dedicated parameter tuning.

\section{Detailed Results on All 11 Datasets of Few-Shot Learning}
\cref{few_shot1,few_shot2} provide detailed comparisons of MMRL and MMRL++ with prior state-of-the-art methods on few-shot learning across 11 datasets. MMRL achieves the highest average performance across all shot settings, with MMRL++ ranking second. Note that the results for MMA are reproduced using its official open-source implementation, as the original paper does not report results for this experiment.

\begin{table*}[t]
\small
\centering
\caption{Comparison of MMRL and MMRL++ with previous state-of-the-art methods on few-shot learning across 11 datasets.}
\label{few_shot1}
\setlength{\tabcolsep}{15pt}{
\resizebox{1.0\textwidth}{!}{
    \begin{tabular}{ll|ccccc}
    \toprule
    \textbf{Dataset} &
      \textbf{Method} &
      \textbf{1 shot} &
      \textbf{2 shots} &
      \textbf{4 shots} &
      \textbf{8 shots} &
      \textbf{16 shots} \\ \midrule
     &
      Linear probe CLIP &
      45.83 &
      57.98 &
      68.01 &
      74.47 &
      78.79 \\
     &
      CoOp &
      67.56 &
      70.65 &
      74.02 &
      76.98 &
      79.89 \\
     &
      CoCoOp &
      66.79 &
      67.65 &
      71.21 &
      72.96 &
      74.90 \\
     &
      MaPLe &
      69.27 &
      72.58 &
      75.37 &
      78.89 &
      81.79 \\
     &
      PromptSRC &
      72.32 &
      75.29 &
      78.35 &
      80.69 &
      82.87 \\
     &
      MMA &
      69.28 &
      72.08 &
      76.38 &
      79.57 &
      82.76 \\
     &
      \cellcolor[HTML]{E8E8E8}$\text{MMRL}_{\text{ (Ours)}}$ &
      \cellcolor[HTML]{E8E8E8}\textbf{72.67} &
      \cellcolor[HTML]{E8E8E8}\textbf{75.90} &
      \cellcolor[HTML]{E8E8E8}\textbf{79.20} &
      \cellcolor[HTML]{E8E8E8}\textbf{81.47} &
      \cellcolor[HTML]{E8E8E8}\textbf{84.34} \\
    \multirow{-8}{*}{Average} &
      \cellcolor[HTML]{E8E8E8}$\text{MMRL++}_{\text{ (Ours)}}$ &
      \cellcolor[HTML]{E8E8E8}72.66 &
      \cellcolor[HTML]{E8E8E8}75.25 &
      \cellcolor[HTML]{E8E8E8}79.00 &
      \cellcolor[HTML]{E8E8E8}81.33 &
      \cellcolor[HTML]{E8E8E8}84.05 \\ \midrule
     &
      Linear probe CLIP &
      32.13 &
      44.88 &
      54.85 &
      62.23 &
      67.31 \\
     &
      CoOp &
      66.33 &
      67.07 &
      68.73 &
      70.63 &
      71.87 \\
     &
      CoCoOp &
      69.43 &
      69.78 &
      70.39 &
      70.63 &
      70.83 \\
     &
      MaPLe &
      62.67 &
      65.10 &
      67.70 &
      70.30 &
      72.33 \\
     &
      PromptSRC &
      68.13 &
      69.77 &
      71.07 &
      \textbf{72.33} &
      73.17 \\
     &
      MMA &
      69.17 &
      70.37 &
      71.00 &
      71.77 &
      73.13 \\
     &
      \cellcolor[HTML]{E8E8E8}$\text{MMRL}_{\text{ (Ours)}}$ &
      \cellcolor[HTML]{E8E8E8}69.00 &
      \cellcolor[HTML]{E8E8E8}70.30 &
      \cellcolor[HTML]{E8E8E8}71.40 &
      \cellcolor[HTML]{E8E8E8}\textbf{72.33} &
      \cellcolor[HTML]{E8E8E8}\textbf{73.40} \\
    \multirow{-8}{*}{ImageNet} &
      \cellcolor[HTML]{E8E8E8}$\text{MMRL++}_{\text{ (Ours)}}$ &
      \cellcolor[HTML]{E8E8E8}\textbf{70.03} &
      \cellcolor[HTML]{E8E8E8}\textbf{70.97} &
      \cellcolor[HTML]{E8E8E8}\textbf{71.47} &
      \cellcolor[HTML]{E8E8E8}72.17 &
      \cellcolor[HTML]{E8E8E8}73.00 \\ \midrule
     &
      Linear probe CLIP &
      79.88 &
      89.01 &
      92.05 &
      93.41 &
      95.43 \\
     &
      CoOp &
      92.60 &
      93.07 &
      94.40 &
      94.37 &
      95.57 \\
     &
      CoCoOp &
      93.83 &
      94.82 &
      94.98 &
      95.04 &
      95.16 \\
     &
      MaPLe &
      92.57 &
      93.97 &
      94.43 &
      95.20 &
      96.00 \\
     &
      PromptSRC &
      93.67 &
      94.53 &
      95.27 &
      95.67 &
      96.07 \\
     &
      MMA &
      92.90 &
      94.00 &
      94.33 &
      95.37 &
      96.33 \\
     &
      \cellcolor[HTML]{E8E8E8}$\text{MMRL}_{\text{ (Ours)}}$ &
      \cellcolor[HTML]{E8E8E8}\textbf{94.17} &
      \cellcolor[HTML]{E8E8E8}94.83 &
      \cellcolor[HTML]{E8E8E8}\textbf{96.03} &
      \cellcolor[HTML]{E8E8E8}\textbf{96.27} &
      \cellcolor[HTML]{E8E8E8}\textbf{97.13} \\
    \multirow{-8}{*}{Caltech101} &
      \cellcolor[HTML]{E8E8E8}$\text{MMRL++}_{\text{ (Ours)}}$ &
      \cellcolor[HTML]{E8E8E8}94.07 &
      \cellcolor[HTML]{E8E8E8}\textbf{94.90} &
      \cellcolor[HTML]{E8E8E8}95.80 &
      \cellcolor[HTML]{E8E8E8}96.13 &
      \cellcolor[HTML]{E8E8E8}96.73 \\ \midrule
     &
      Linear probe CLIP &
      44.06 &
      58.37 &
      71.17 &
      78.36 &
      85.34 \\
     &
      CoOp &
      90.37 &
      89.80 &
      92.57 &
      91.27 &
      91.87 \\
     &
      CoCoOp &
      91.27 &
      \textbf{92.64} &
      92.81 &
      93.45 &
      93.34 \\
     &
      MaPLe &
      89.10 &
      90.87 &
      91.90 &
      92.57 &
      92.83 \\
     &
      PromptSRC &
      \textbf{92.00} &
      92.50 &
      \textbf{93.43} &
      \textbf{93.50} &
      93.67 \\
     &
      MMA &
      91.23 &
      91.97 &
      92.23 &
      92.77 &
      93.23 \\
     &
      \cellcolor[HTML]{E8E8E8}$\text{MMRL}_{\text{ (Ours)}}$ &
      \cellcolor[HTML]{E8E8E8}90.87 &
      \cellcolor[HTML]{E8E8E8}91.57 &
      \cellcolor[HTML]{E8E8E8}92.57 &
      \cellcolor[HTML]{E8E8E8}93.03 &
      \cellcolor[HTML]{E8E8E8}\textbf{93.83} \\
    \multirow{-8}{*}{OxfordPets} &
      \cellcolor[HTML]{E8E8E8}$\text{MMRL++}_{\text{ (Ours)}}$ &
      \cellcolor[HTML]{E8E8E8}90.97 &
      \cellcolor[HTML]{E8E8E8}91.17 &
      \cellcolor[HTML]{E8E8E8}92.43 &
      \cellcolor[HTML]{E8E8E8}92.70 &
      \cellcolor[HTML]{E8E8E8}93.30 \\ \midrule
     &
      Linear probe CLIP &
      35.66 &
      50.28 &
      63.38 &
      73.67 &
      80.44 \\
     &
      CoOp &
      67.43 &
      70.50 &
      74.47 &
      79.30 &
      83.07 \\
     &
      CoCoOp &
      67.22 &
      68.37 &
      69.39 &
      70.44 &
      71.57 \\
     &
      MaPLe &
      66.60 &
      71.60 &
      75.30 &
      79.47 &
      83.57 \\
     &
      PromptSRC &
      \textbf{69.40} &
      \textbf{73.40} &
      77.13 &
      80.97 &
      83.83 \\
     &
      MMA &
      67.87 &
      71.77 &
      76.50 &
      81.40 &
      85.70 \\
     &
      \cellcolor[HTML]{E8E8E8}$\text{MMRL}_{\text{ (Ours)}}$ &
      \cellcolor[HTML]{E8E8E8}68.70 &
      \cellcolor[HTML]{E8E8E8}72.93 &
      \cellcolor[HTML]{E8E8E8}\textbf{78.17} &
      \cellcolor[HTML]{E8E8E8}\textbf{82.57} &
      \cellcolor[HTML]{E8E8E8}\textbf{86.43} \\
    \multirow{-8}{*}{StanfordCars} &
      \cellcolor[HTML]{E8E8E8}$\text{MMRL++}_{\text{ (Ours)}}$ &
      \cellcolor[HTML]{E8E8E8}68.20 &
      \cellcolor[HTML]{E8E8E8}72.70 &
      \cellcolor[HTML]{E8E8E8}77.83 &
      \cellcolor[HTML]{E8E8E8}82.40 &
      \cellcolor[HTML]{E8E8E8}86.20 \\ \midrule
     &
      Linear probe CLIP &
      69.74 &
      85.07 &
      92.02 &
      96.10 &
      97.37 \\
     &
      CoOp &
      77.53 &
      87.33 &
      92.17 &
      94.97 &
      97.07 \\
     &
      CoCoOp &
      72.08 &
      75.79 &
      78.40 &
      84.30 &
      87.84 \\
     &
      MaPLe &
      83.30 &
      88.93 &
      92.67 &
      95.80 &
      97.00 \\
     &
      PromptSRC &
      85.93 &
      91.17 &
      93.87 &
      96.27 &
      97.60 \\
     &
      MMA &
      83.60 &
      90.30 &
      93.00 &
      95.97 &
      97.97 \\
     &
      \cellcolor[HTML]{E8E8E8}$\text{MMRL}_{\text{ (Ours)}}$ &
      \cellcolor[HTML]{E8E8E8}\textbf{85.97} &
      \cellcolor[HTML]{E8E8E8}\textbf{91.20} &
      \cellcolor[HTML]{E8E8E8}\textbf{94.60} &
      \cellcolor[HTML]{E8E8E8}\textbf{96.60} &
      \cellcolor[HTML]{E8E8E8}\textbf{98.40} \\
    \multirow{-8}{*}{Flowers102} &
      \cellcolor[HTML]{E8E8E8}$\text{MMRL++}_{\text{ (Ours)}}$ &
      \cellcolor[HTML]{E8E8E8}84.13 &
      \cellcolor[HTML]{E8E8E8}89.70 &
      \cellcolor[HTML]{E8E8E8}93.60 &
      \cellcolor[HTML]{E8E8E8}96.33 &
      \cellcolor[HTML]{E8E8E8}98.20 \\ \bottomrule
    \end{tabular}
    }
    }
\end{table*}

\begin{table*}[t]
\small
\centering
\caption{Comparison of MMRL and MMRL++ with previous state-of-the-art methods on few-shot learning across 11 datasets.}
\label{few_shot2}
\setlength{\tabcolsep}{15pt}{
\resizebox{1.0\textwidth}{!}{
    \begin{tabular}{llccccc}
    \toprule
    \textbf{Dataset} &
      \multicolumn{1}{l|}{\textbf{Method}} &
      \textbf{1 shot} &
      \textbf{2 shots} &
      \textbf{4 shots} &
      \textbf{8 shots} &
      \textbf{16 shots} \\ \midrule
     &
      \multicolumn{1}{l|}{Linear probe CLIP} &
      43.96 &
      61.51 &
      73.19 &
      79.79 &
      82.90 \\
     &
      \multicolumn{1}{l|}{CoOp} &
      84.33 &
      84.40 &
      84.47 &
      82.67 &
      84.20 \\
     &
      \multicolumn{1}{l|}{CoCoOp} &
      \textbf{85.65} &
      \textbf{86.22} &
      \textbf{86.88} &
      \textbf{86.97} &
      87.25 \\
     &
      \multicolumn{1}{l|}{MaPLe} &
      80.50 &
      81.47 &
      81.77 &
      83.60 &
      85.33 \\
     &
      \multicolumn{1}{l|}{PromptSRC} &
      84.87 &
      85.70 &
      86.17 &
      86.90 &
      \textbf{87.50} \\
     &
      \multicolumn{1}{l|}{MMA} &
      83.03 &
      82.50 &
      82.13 &
      83.00 &
      84.57 \\
     &
      \cellcolor[HTML]{E8E8E8}$\text{MMRL}_{\text{ (Ours)}}$ &
      \cellcolor[HTML]{E8E8E8}84.87 &
      \cellcolor[HTML]{E8E8E8}85.53 &
      \cellcolor[HTML]{E8E8E8}85.77 &
      \cellcolor[HTML]{E8E8E8}86.33 &
      \cellcolor[HTML]{E8E8E8}87.03 \\
    \multirow{-8}{*}{Food101} &
      \multicolumn{1}{l|}{\cellcolor[HTML]{E8E8E8}$\text{MMRL++}_{\text{ (Ours)}}$} &
      \cellcolor[HTML]{E8E8E8}83.23 &
      \cellcolor[HTML]{E8E8E8}84.07 &
      \cellcolor[HTML]{E8E8E8}84.83 &
      \cellcolor[HTML]{E8E8E8}85.57 &
      \cellcolor[HTML]{E8E8E8}86.07 \\ \midrule
     &
      \multicolumn{1}{l|}{Linear probe CLIP} &
      19.61 &
      26.41 &
      32.33 &
      39.35 &
      45.36 \\
     &
      \multicolumn{1}{l|}{CoOp} &
      21.37 &
      26.20 &
      30.83 &
      39.00 &
      43.40 \\
     &
      \multicolumn{1}{l|}{CoCoOp} &
      12.68 &
      15.06 &
      24.79 &
      26.61 &
      31.21 \\
     &
      \multicolumn{1}{l|}{MaPLe} &
      26.73 &
      30.90 &
      34.87 &
      42.00 &
      48.40 \\
     &
      \multicolumn{1}{l|}{PromptSRC} &
      27.67 &
      31.70 &
      37.47 &
      43.27 &
      50.83 \\
     &
      \multicolumn{1}{l|}{MMA} &
      \textbf{28.73} &
      31.90 &
      37.57 &
      44.83 &
      52.70 \\
     &
      \cellcolor[HTML]{E8E8E8}$\text{MMRL}_{\text{ (Ours)}}$ &
      \cellcolor[HTML]{E8E8E8}28.53 &
      \cellcolor[HTML]{E8E8E8}\textbf{34.23} &
      \cellcolor[HTML]{E8E8E8}40.47 &
      \cellcolor[HTML]{E8E8E8}48.07 &
      \cellcolor[HTML]{E8E8E8}57.60 \\
    \multirow{-8}{*}{FGVCAircraft} &
      \multicolumn{1}{l|}{\cellcolor[HTML]{E8E8E8}$\text{MMRL++}_{\text{ (Ours)}}$} &
      \cellcolor[HTML]{E8E8E8}28.10 &
      \cellcolor[HTML]{E8E8E8}32.40 &
      \cellcolor[HTML]{E8E8E8}\textbf{40.87} &
      \cellcolor[HTML]{E8E8E8}\textbf{49.23} &
      \cellcolor[HTML]{E8E8E8}\textbf{58.20} \\ \midrule
     &
      \multicolumn{1}{l|}{Linear probe CLIP} &
      41.58 &
      53.70 &
      63.00 &
      69.08 &
      73.28 \\
     &
      \multicolumn{1}{l|}{CoOp} &
      66.77 &
      66.53 &
      69.97 &
      71.53 &
      74.67 \\
     &
      \multicolumn{1}{l|}{CoCoOp} &
      68.33 &
      69.03 &
      70.21 &
      70.84 &
      72.15 \\
     &
      \multicolumn{1}{l|}{MaPLe} &
      64.77 &
      67.10 &
      70.67 &
      73.23 &
      75.53 \\
     &
      \multicolumn{1}{l|}{PromptSRC} &
      \textbf{69.67} &
      \textbf{71.60} &
      \textbf{74.00} &
      75.73 &
      77.23 \\
     &
      \multicolumn{1}{l|}{MMA} &
      64.00 &
      67.17 &
      69.97 &
      72.30 &
      74.63 \\
     &
      \cellcolor[HTML]{E8E8E8}$\text{MMRL}_{\text{ (Ours)}}$ &
      \cellcolor[HTML]{E8E8E8}68.90 &
      \cellcolor[HTML]{E8E8E8}71.53 &
      \cellcolor[HTML]{E8E8E8}73.93 &
      \cellcolor[HTML]{E8E8E8}\textbf{76.00} &
      \cellcolor[HTML]{E8E8E8}\textbf{77.70} \\
    \multirow{-8}{*}{SUN397} &
      \multicolumn{1}{l|}{\cellcolor[HTML]{E8E8E8}$\text{MMRL++}_{\text{ (Ours)}}$} &
      \cellcolor[HTML]{E8E8E8}68.97 &
      \cellcolor[HTML]{E8E8E8}70.97 &
      \cellcolor[HTML]{E8E8E8}73.43 &
      \cellcolor[HTML]{E8E8E8}75.53 &
      \cellcolor[HTML]{E8E8E8}77.50 \\ \midrule
     &
      \multicolumn{1}{l|}{Linear probe CLIP} &
      34.59 &
      40.76 &
      55.71 &
      63.46 &
      69.96 \\
     &
      \multicolumn{1}{l|}{CoOp} &
      50.23 &
      53.60 &
      58.70 &
      64.77 &
      69.87 \\
     &
      \multicolumn{1}{l|}{CoCoOp} &
      48.54 &
      52.17 &
      55.04 &
      58.89 &
      63.04 \\
     &
      \multicolumn{1}{l|}{MaPLe} &
      52.13 &
      55.50 &
      61.00 &
      66.50 &
      71.33 \\
     &
      \multicolumn{1}{l|}{PromptSRC} &
      56.23 &
      59.97 &
      65.53 &
      69.87 &
      72.73 \\
     &
      \multicolumn{1}{l|}{MMA} &
      52.27 &
      56.90 &
      63.93 &
      67.97 &
      73.47 \\
     &
      \cellcolor[HTML]{E8E8E8}$\text{MMRL}_{\text{ (Ours)}}$ &
      \cellcolor[HTML]{E8E8E8}56.37 &
      \cellcolor[HTML]{E8E8E8}\textbf{61.37} &
      \cellcolor[HTML]{E8E8E8}\textbf{67.87} &
      \cellcolor[HTML]{E8E8E8}\textbf{71.60} &
      \cellcolor[HTML]{E8E8E8}\textbf{75.30} \\
    \multirow{-8}{*}{DTD} &
      \multicolumn{1}{l|}{\cellcolor[HTML]{E8E8E8}$\text{MMRL++}_{\text{ (Ours)}}$} &
      \cellcolor[HTML]{E8E8E8}\textbf{57.00} &
      \cellcolor[HTML]{E8E8E8}60.80 &
      \cellcolor[HTML]{E8E8E8}67.03 &
      \cellcolor[HTML]{E8E8E8}70.83 &
      \cellcolor[HTML]{E8E8E8}74.37 \\ \midrule
     &
      \multicolumn{1}{l|}{Linear probe CLIP} &
      49.23 &
      61.98 &
      77.09 &
      84.43 &
      87.21 \\
     &
      \multicolumn{1}{l|}{CoOp} &
      54.93 &
      65.17 &
      70.80 &
      78.07 &
      84.93 \\
     &
      \multicolumn{1}{l|}{CoCoOp} &
      55.33 &
      46.74 &
      65.56 &
      68.21 &
      73.32 \\
     &
      \multicolumn{1}{l|}{MaPLe} &
      71.80 &
      78.30 &
      84.50 &
      87.73 &
      92.33 \\
     &
      \multicolumn{1}{l|}{PromptSRC} &
      73.13 &
      79.37 &
      86.30 &
      88.80 &
      92.43 \\
     &
      \multicolumn{1}{l|}{MMA} &
      55.07 &
      59.80 &
      79.40 &
      86.47 &
      92.37 \\
     &
      \cellcolor[HTML]{E8E8E8}$\text{MMRL}_{\text{ (Ours)}}$ &
      \cellcolor[HTML]{E8E8E8}76.00 &
      \cellcolor[HTML]{E8E8E8}\textbf{82.87} &
      \cellcolor[HTML]{E8E8E8}87.67 &
      \cellcolor[HTML]{E8E8E8}88.73 &
      \cellcolor[HTML]{E8E8E8}93.37 \\
    \multirow{-8}{*}{EuroSAT} &
      \multicolumn{1}{l|}{\cellcolor[HTML]{E8E8E8}$\text{MMRL++}_{\text{ (Ours)}}$} &
      \cellcolor[HTML]{E8E8E8}\textbf{79.07} &
      \cellcolor[HTML]{E8E8E8}81.17 &
      \cellcolor[HTML]{E8E8E8}\textbf{89.23} &
      \cellcolor[HTML]{E8E8E8}\textbf{89.17} &
      \cellcolor[HTML]{E8E8E8}\textbf{93.50} \\ \midrule
     &
      \multicolumn{1}{l|}{Linear probe CLIP} &
      53.66 &
      65.78 &
      73.28 &
      79.34 &
      82.11 \\
     &
      \multicolumn{1}{l|}{CoOp} &
      71.23 &
      73.43 &
      77.10 &
      80.20 &
      82.23 \\
     &
      \multicolumn{1}{l|}{CoCoOp} &
      70.30 &
      73.51 &
      74.82 &
      77.14 &
      78.14 \\
     &
      \multicolumn{1}{l|}{MaPLe} &
      71.83 &
      74.60 &
      78.47 &
      81.37 &
      85.03 \\
     &
      \multicolumn{1}{l|}{PromptSRC} &
      74.80 &
      78.50 &
      81.57 &
      84.30 &
      86.47 \\
     &
      \multicolumn{1}{l|}{MMA} &
      74.17 &
      76.17 &
      80.10 &
      83.43 &
      86.30 \\
     &
      \cellcolor[HTML]{E8E8E8}$\text{MMRL}_{\text{ (Ours)}}$ &
      \cellcolor[HTML]{E8E8E8}\textbf{75.97} &
      \cellcolor[HTML]{E8E8E8}78.50 &
      \cellcolor[HTML]{E8E8E8}\textbf{82.67} &
      \cellcolor[HTML]{E8E8E8}\textbf{84.67} &
      \cellcolor[HTML]{E8E8E8}\textbf{87.60} \\
    \multirow{-8}{*}{UCF101} &
      \multicolumn{1}{l|}{\cellcolor[HTML]{E8E8E8}$\text{MMRL++}_{\text{ (Ours)}}$} &
      \cellcolor[HTML]{E8E8E8}75.50 &
      \cellcolor[HTML]{E8E8E8}\textbf{78.87} &
      \cellcolor[HTML]{E8E8E8}82.50 &
      \cellcolor[HTML]{E8E8E8}84.53 &
      \cellcolor[HTML]{E8E8E8}87.43 \\ \bottomrule
    \end{tabular}
    }
    }
\end{table*}

\end{appendices}

\end{document}